\documentclass[10pt,journal,compsoc,twoside]{IEEEtran}

\usepackage{amsmath,amsfonts}
\usepackage{algorithmic}
\usepackage{algorithm}
\usepackage{array}
\usepackage[nocompress]{cite}
\usepackage[caption=false,font=normalsize,labelfont=sf,textfont=sf]{subfig}
\usepackage{textcomp}
\usepackage{stfloats}
\usepackage{url}
\usepackage{verbatim}
\usepackage{amsmath,amsfonts}
\usepackage{microtype}
\usepackage{booktabs}
\usepackage{balance}

\usepackage[pdftex, bookmarksnumbered=true, pagebackref=false]{hyperref}
\hypersetup{colorlinks,
	citecolor=[rgb]{0.0,0.0,0.0},
	filecolor=[rgb]{0.6,0.0,0.4},
	linkcolor=[rgb]{0.0,0.0,0.0},
	urlcolor=[rgb]{0.0,0.0,0.4}
}

\usepackage{paralist}

\usepackage{graphicx}
\usepackage{cite}
\usepackage{threeparttable}
\usepackage{makecell}
\usepackage{multirow}
\usepackage{footnote}
\usepackage{tablefootnote}
\usepackage{subfig,float}
\usepackage[numbers]{natbib}
\usepackage{colortbl}
\usepackage{array}
\usepackage{caption}
\usepackage{booktabs}

\begin{document}

\title{Privacy-Preserving Biometric Verification with Handwritten Random Digit String}

\author{Peirong Zhang,
		Yuliang Liu,
		Songxuan Lai,
		Hongliang Li,
		Lianwen Jin$^*$~\IEEEmembership{Member,~IEEE}

\IEEEcompsocitemizethanks{
\IEEEcompsocthanksitem Peirong Zhang, Hongliang Li, and Lianwen Jin are with the School of Electronic and Information Engineering, South China University of Technology, Guangzhou, China. Lianwen Jin is also with SCUT-Zhuhai Institute of Modern Industrial Innovation, Zhuhai, China.
\IEEEcompsocthanksitem Yuliang Liu is with the School of Artificial Intelligence and Automation, Huazhong University of Science and Technology, Wuhan, China.
\IEEEcompsocthanksitem Songxuan Lai is with the EI Product Department, Huawei Cloud Computing Technologies Co., Ltd., Shenzhen, China. This work was done while pursuing his Ph.D. at the South China University of Technology.}
\thanks{$^*$ Corresponding author.}
}

\markboth
{IEEE Transactions on Pattern Analysis and Machine Intelligence, VOL.~47, NO.~4, APRIL~2025}
{Zhang \MakeLowercase{et al.}: Privacy-Preserving Biometric Verification with Handwritten Random Digit String}

\IEEEtitleabstractindextext{
\begin{abstract}
Handwriting verification has stood as a steadfast identity authentication method for decades. However, this technique risks potential privacy breaches due to the inclusion of personal information in handwritten biometrics such as signatures. To address this concern, we propose using the Random Digit String (RDS) for privacy-preserving handwriting verification. This approach allows users to authenticate themselves by writing an arbitrary digit sequence, effectively ensuring privacy protection. To evaluate the effectiveness of RDS, we construct a new HRDS4BV dataset composed of online naturally handwritten RDS. Unlike conventional handwriting, RDS encompasses unconstrained and variable content, posing significant challenges for modeling consistent personal writing style. To surmount this, we propose the Pattern Attentive VErification Network (PAVENet), along with a Discriminative Pattern Mining (DPM) module. DPM adaptively enhances the recognition of consistent and discriminative writing patterns, thus refining handwriting style representation. Through comprehensive evaluations, we scrutinize the applicability of online RDS verification and showcase a pronounced outperformance of our model over existing methods. Furthermore, we discover a noteworthy forgery phenomenon that deviates from prior findings and discuss its positive impact in countering malicious impostor attacks. Substantially, our work underscores the feasibility of privacy-preserving biometric verification and propels the prospects of its broader acceptance and application.
\end{abstract}

}

\maketitle

\IEEEdisplaynontitleabstractindextext

\IEEEraisesectionheading{\section{Introduction}\label{sec:introduction}}

\IEEEPARstart{H}{andwriting} identity verification is to authenticate personal identity through distinguishing genuine and forged handwriting samples (\emph{e.g.}, signatures). It is of high societal acceptance in modern e-society as a practical means for information security, which is widely used in forensics, contracts, last wills, and financial transactions \cite{jain2004anintro,diaz2019aperspective,lai2021synsig2vec}. Handwriting verification techniques can be typically categorized as online and offline methods \cite{svd2008kamel,plamondon1989automatic} based on data acquisition manners. Online verification methods utilize dynamic handwriting information such as temporal and positional clues produced in the writing process, whereas the offline counterparts rely on static images captured by scanners or cameras. This study targets online handwriting verification. Generally, handwriting verification systems are required to contend with two types of forgery attacks: skilled forgery and random forgery. Skilled forgery refers to the intentional imitation of genuine handwriting produced by skilled and experienced forgers, aiming to closely resemble the original. In contrast, random forgery involves handwriting created without prior knowledge of the the imitated writer's writing style. To fortify its defenses, a robust verification system is typically trained with a combination of genuine data and both types of forgery data. This training process enhances the system's ability to discern the truth, enabling it to accurately determine the authenticity of a query handwriting.

In the spurt of the Internet and AI, privacy breaches have become increasingly common, heightening public awareness and concern regarding privacy security. Considerable exertions have been dedicated to preserving privacy in scenarios involving privacy-sensitive data, such as medical \cite{kaissis2020secure,bai2021advancing,yang2022digital}, Internet of Things (IoT) and blockchains \cite{iotoverview2020huo,kosba2016blockchain,lu2020blockchain}, cloud system application \cite{clouddata2014cao,cloudcompute2016xia}, and social network service \cite{negihborattack2008zhou,customized2018he,wu2022federated}. However, while handwriting verification aims to facilitate identity recognition, it harbors long-standing and unresolved privacy leakage issues. A prime example is online signature verification, the most popular handwriting verification scheme, where personally identifiable information within signatures can potentially result in privacy leakage, raising user apprehensions. The privacy concern not only undermines the reliability of handwriting verification systems in applications, but also impedes large-scale handwriting data collection and therefore the development of robust verification systems that demand substantial data. Recently, researchers have explored using separate digits or consecutive digit strings in online handwriting verification \cite{zhang2022msds,tolosana2019biotouchpass,tolosana2020biotouchpass2} for their distinctive writing features, such as the Token Digit String (TDS) \cite{zhang2022msds}. However, akin to the signature, TDS also encompasses identifiable phone numbers, raising similar privacy concerns and confronting the same data collection constraint.

Fundamentally, the private information within signatures/TDS arises from their content, \emph{i.e.}, the name/phone number. In contrast, when the lexical content of the text is arbitrary, it becomes privacy-free and therefore privacy-preserving. To be more specific, handwriting with arbitrary content is defined as text-independent (TI) \cite{campbell1997speaker} handwriting, such as poems and merchandise numbers, whereas handwriting with fixed content is defined as text-dependent (TD) \cite{kinnunen2010overiew}, such as signature and TDS. Due to its privacy-free intrinsic, TI handwriting offers a promising privacy-preserving alternative to the privacy-sensitive signature/TDS. Nonetheless, despite its potential, no previous attempts have explored TI handwriting verification. Suitable datasets for training and testing such systems are not yet available, either. To clarify, there is a similar field to handwriting verification known as writer identification \cite{yang2016deepwriterid,zhang2017end2end,rehman2019writer}, which typically involves TI scenarios. However, the data from this domain can not be directly adapted to the verification task owing to the absence of corresponding skilled forgery of genuine data, which is essential for training verification models.

\begin{figure}[t]
	\centering
	\includegraphics[width=0.88\linewidth]{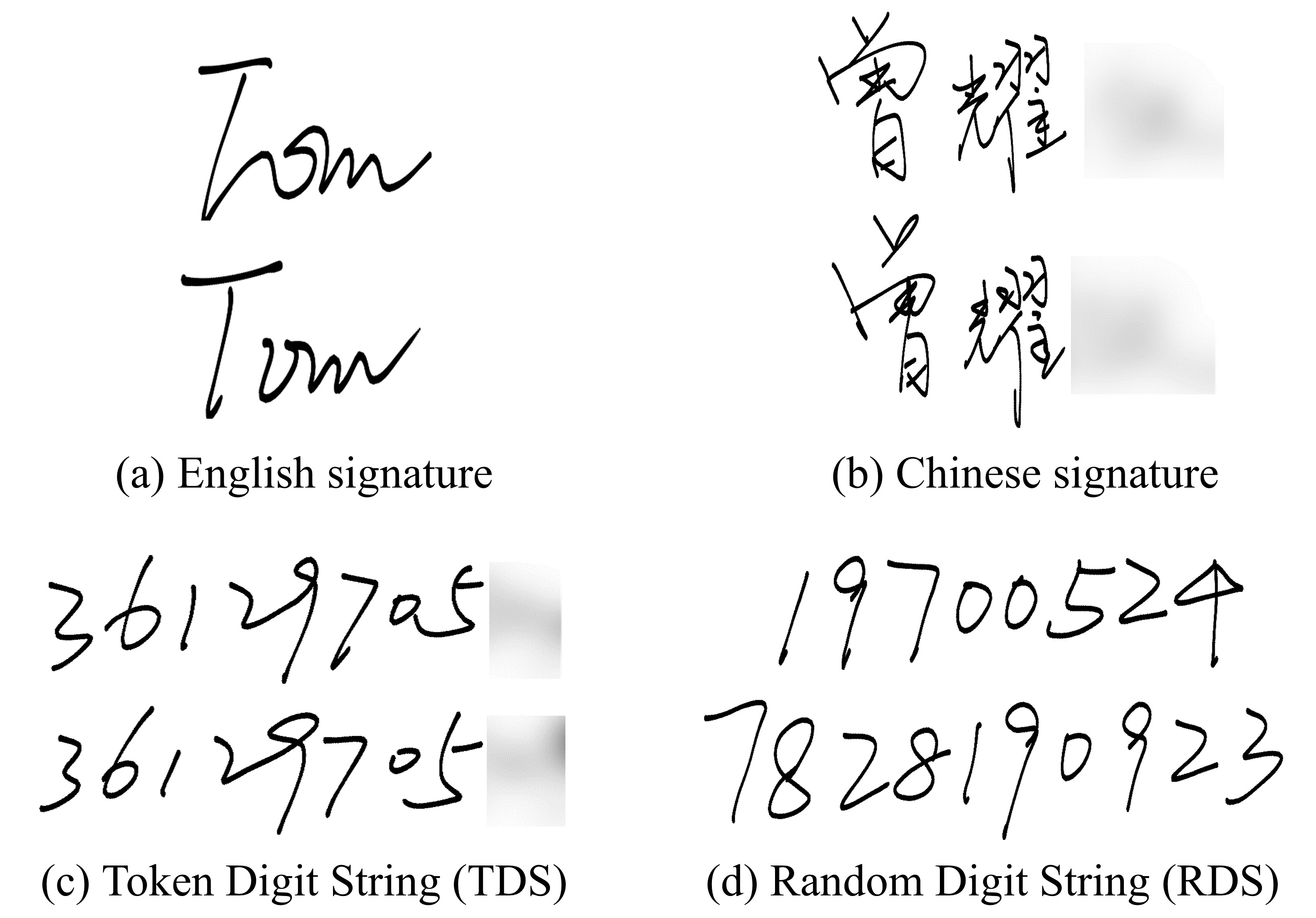}{}
	\caption{Visualization of our proposed online handwritten RDS and comparisons with signatures and TDS. The samples of Chinese signature and TDS are obtained from the MSDS dataset \cite{zhang2022msds}, which are desensitized through mosaic to protect individual privacy.}
	\label{Fig::comparison}
\end{figure}

In this paper, we propose to use the natural-written string composed of random digits, \emph{a.k.a.}, Random Digit String (RDS), as a new biometric trait for online handwriting verification. Our motivation involves two key aspects. First, the primary merit of using RDS for biometric verification lies in its inherent privacy preservation nature that stems from its text independence. RDS is random and arbitrary in terms of content composition, devoid of any risk of privacy exposure or disclosure. An individual can simply write an arbitrary digit sequence instead of signing a signature to attain authentication without divulging any private data. This unique privacy feature sets RDS apart from other biometrics and serves as a pivotal driver for adopting RDS in our pursuit of privacy-preserving handwriting verification. Second, RDS is a ubiquitous occurrence in daily life, appearing in various contexts such as date strings, contract numbers, and order identifiers. People are familiar with writing these sequences of Arabic numerals in a natural and consecutive manner, leading to the development of strong muscle memory. Consequently, naturally handwritten RDS manifests a rich reservoir of discriminative writing characteristics. Each individual imparts their unique writing style to the handwritten RDS, endowing it with distinctiveness and reliability as a biometric identifier. These two advantages render RDS a promising and valuable tool for privacy-preserving verification.

To assess the applicability and effectiveness of RDS in online biometric verification, we build a new Handwritten Random Digit String for Biometric Verification (HRDS4BV) dataset, which comprises online naturally handwritten RDS collected from 402 writers. We collect genuine handwriting samples of writers and corresponding skilled forged samples in an equal number, with 20 genuine/forged ones in two captured sessions, resulting in 8,040 genuine samples and skilled forgeries, respectively. Several RDS examples are visualized in Fig.~\ref{Fig::comparison}.

\begin{figure}[t]
	\centering
	\includegraphics[width=\linewidth]{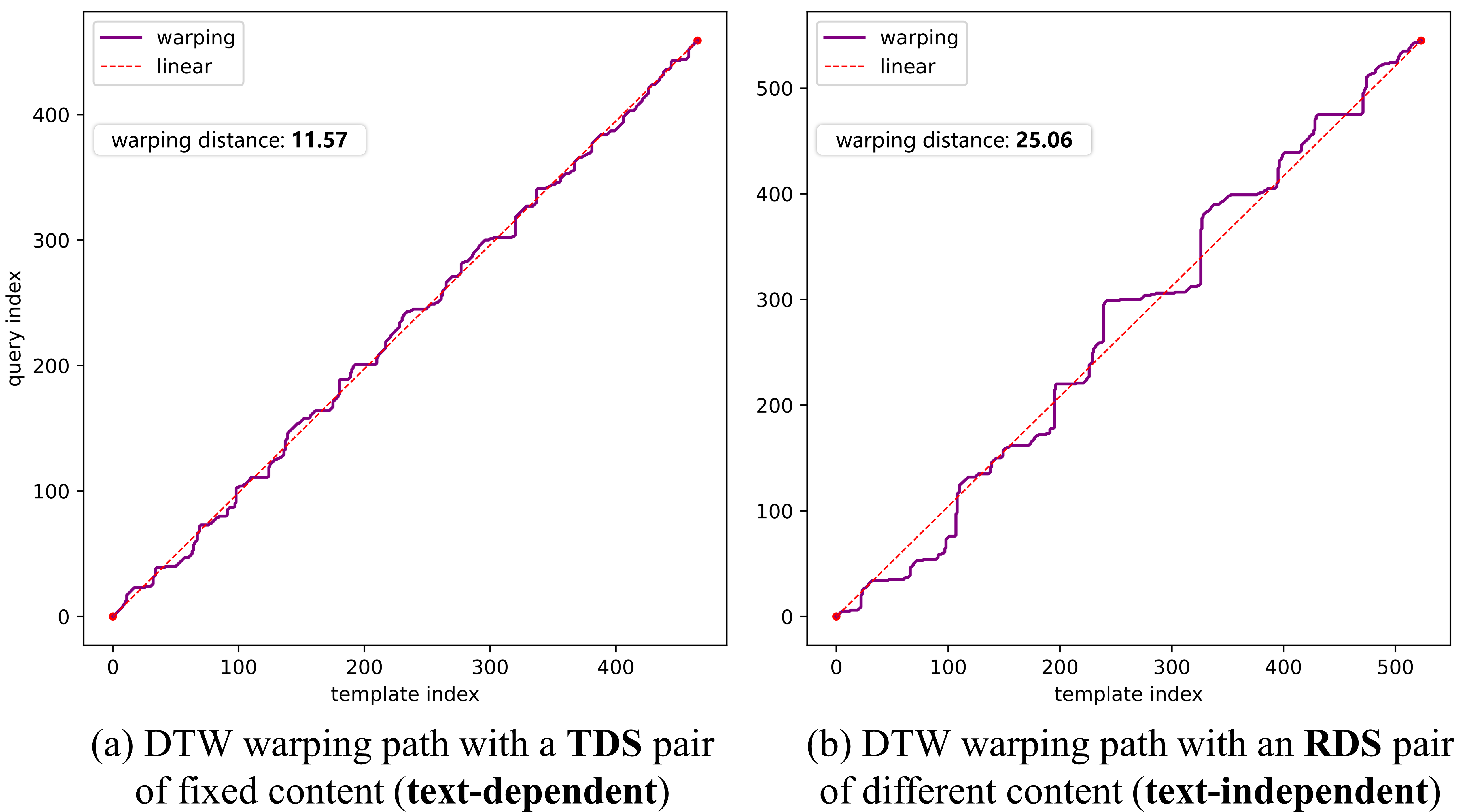}{}
	\caption{Warping paths of DTW between different query and template pairs. The left results are with a TDS pair of fixed content, whereas the right ones are with an RDS pair of different content. Smoother curve and smaller warping distance are better.}
	\label{Fig::dtw}
\end{figure}

Although online RDS verification offers an exicting avenue for privacy-preserving identity verification, it grapples with challenges in handling text-independent handwriting with highly variable content. In TD scenarios, writers are confined to writing particular and fixed content, such as signature/TDS, showcasing stable writing styles. In TI scenarios, however, writers exhibit diverse character formation across variable content, resulting in significant writing style variations. This poses several challenges for TI handwriting verification, including (1) higher intra-class variation (variation within the same writer), (2) potential obfuscation of distinct styles, and (3) difficulties posed by text complexity, such as varying content lengths. Consequently, TD verification systems can easily learn fixed content traits and consistent writing features, whereas TI systems struggle with establishing robust writing style representation across diverse and unconstrained content.

So far, while existing methods \cite{vintsyuk1968speech,lai2018recurrent,lai2021synsig2vec,jiang2022dsdtw,wu2019DeepDTW,tolosana2020biotouchpass2} tailored for TD handwriting verification have made remarkable strides, their application to TI scenarios might be hindered by two issues. (1) Models may fail to form smooth writing feature representations in the absence of fixed content. For example, when we employ the well-known Dynamic Time Warping (DTW) \cite{vintsyuk1968speech} on a text-independent RDS pair and a text-dependent TDS pair, the results reveal a stark contrast. As shown in Fig.~\ref{Fig::dtw}, given RDS data, the warping path appears significantly more rugged and the warping distance is much larger. (2) Without the guidance of fixed content in TD scenarios, models could struggle to recognize the unique personal writing style across a wide range of text content. This limitation significantly damages the verification accuracy. However, the majority of prior models heavily rely on the content information and understand the writing features in a coarse manner, neglecting the excavation of distinctive styles and writing patterns. The deficiency in style modeling hampers the establishment of consistent writing characteristic representation for handwriting with variable content such as RDS, which is particularly crucial for verification. Compelled by the above observations, existing methods can not be directly transplanted into RDS verification while meeting satisfactory performances.

To this end, we propose a Pattern Attentive VErificaton Network (PAVENet) to handle RDS verification in unconstrained writing scenarios, drawing inspiration from human writing habits and visual perception of handwritten text. Over time, individuals develop exclusive and recurring writing patterns due to their long-established writing habits, such as hyphenation, stroke twirls, and stroke flourishes. These salient handwriting patterns are consistent and discriminative, making them potential key verification clues if properly highlighted. Consequently, we design a Discriminative Pattern Mining (DPM) module to excavate these discriminative patterns in personal handwriting. DPM adaptively attends to various patterns from a high-dimensional perspective, which effectively empowers the model's style modeling ability that has been previously disregarded. This enhancement substantially optimizes the representations of unconstrained handwriting styles in the TI context. To assess the applicability of RDS verification and validate the effectiveness of our proposed model, we conduct extensive experiments on the HRDS4BV and the e-BioDigit \cite{tolosana2019biotouchpass} datasets. The results demonstrate that our proposed method significantly outperforms preceding approaches in multiple aspects, substantiating its superiority in RDS verification. In addition, we reveal an intriguing phenomenon that skilled forgery is relatively easier to verify than random forgery in RDS verification, which contrasts with previous studies. This observation further underscores the feasibility of RDS verification and could inspire the design of new strategies in the defense of skilled forgery attacks in practical applications.

To summarize, our major contributions are as follows:
\begin{itemize}
	\item We propose to leverage Random Digit String (RDS) as a new biometric trait for privacy-preserving handwriting verification in an effective and convenient manner. Users can simply write arbitrary digit strings to verify themselves without compromising their privacy. To the best of our knowledge, this is the first attempt that explicitly targets privacy-preserving handwriting verification.
	\item To effectively surmount the challenges in writing style representation posed by the inherent content variability and complexity of RDS, we propose a verification model termed PAVENet, along with a specific DPM module for handwriting pattern mining. Inspired by human writing habits and the visual perception of handwritten text, DPM adaptively attends to consistent and discriminative writing patterns, such as hyphenated strokes, stroke twirls, and stroke flourishes. This targeted attention to key handwriting patterns significantly benefits style learning and facilitates handwriting distinguishing.
	\item We construct a corresponding HRDS4BV dataset for RDS benchmark evaluation. Notably, this dataset is the first of its kind in handwriting verification, being text-independent (TI) and containing coherently and naturally handwritten online RDS samples. The code and HRDS4BV dataset are available at \url{https://github.com/SCUT-DLVCLab/PAVENet}.
	\item Extensive experiments illustrate the effectiveness of employing RDS for handwriting verification and thus underscore the feasibility of privacy-preserving biometric verification. The results demonstrate that our PAVENet significantly outperforms prevailing methods and substantiates the effectiveness of the proposed DPM in extracting natural writing patterns across diverse contents. Notably, skilled forgery is easier verified than random forgery in RDS verification, which differs from previous findings. We delve into a deeper analysis of this phenomenon and deliberate on its potential positive impact in countering malicious impostor attacks.
\end{itemize}

\section{Related Work}
\label{sec::related}
\begin{table*}[ht]
	\renewcommand{\arraystretch}{1.1}
	\caption{Details of the proposed HRDS4BV dataset and comparisons with existing datasets.}
	\label{Table::comparison}
	\centering
	\resizebox{\textwidth}{!}{
	\begin{threeparttable}
		\begin{tabular}{c c c c c c c c c}
			\toprule
			\noalign{\vspace{-2pt}}
			\textbf{Dataset} & \textbf{Public} & \textbf{Content} & \textbf{User} & \textbf{Session} & \textbf{Session Gap} & \textbf{Genuine Sample} & \textbf{Skilled Forgery} & \textbf{Features}\tnote{1}\\
			\hline
			LMCA \cite{el2008off} & \checkmark & Digit \tnote{2} & 55 & 1 & - & 30,000 & - & $X$,$Y$\\
			AOD \cite{arabicdigit2012azeem} & \checkmark & Digit \tnote{2} & 55 & 1 & - & 30,000 & - & $X$,$Y$\\
			MAYASTROUN \cite{multidigit2012njah} & \checkmark & Digit \tnote{2} & 355 & 1 & - & 6,000 & - & $X$,$Y$\\
			Dataset in \cite{nguyen2017draw} & $\times$ & 4-digit PIN & 45 & - & - & 3,203 & 4,655 & $X$,$Y$,$P$,$S$\\
			e-BioDigit \cite{tolosana2019biotouchpass} & \checkmark & Digit & 93 &  2 & $\ge$3 weeks & 7,440 & - & $X$,$Y$,$P$,$T$\\
			MobileTouchDB \cite{tolosana2020biotouchpass2} & \checkmark & Character \& Digit & 217 & 6 & $\ge$2 days & 64K & - & $X$,$Y$\\
			MSDS-TDS \cite{zhang2022msds} & \checkmark & Token Digit String & 402 & 2 & $\ge$3 weeks & 402$\times$20 = 8,040 &  402$\times$20 = 8,040 & $X$,$Y$,$P$,$T$,$I_r$\\
			\hline
			HRDS4BV (\textbf{Ours}) & \checkmark & Random Digit String & 402 & 2 & $\ge$3 weeks & 402$\times$20 = 8,040 & 402$\times$20 = 8,040 & $X$,$Y$,$P$,$T$,$U$,$D$\\
			\noalign{\vspace{-2pt}}
			\bottomrule
		\end{tabular}
		\begin{tablenotes}
			\item[1] $X,Y,P,T,I_r,S,U,D$ respectively denote the $x$ coordinates, $y$ coordinates, pressure, timestamps, rendered images, size of touch (how big the finger is), pen-ups, and pen-downs.
			\item[2] As they are not created for online handwriting verification, we only compare their subset of digits.
		\end{tablenotes}
	\end{threeparttable}}
\end{table*}

\subsection{Privacy Preservation}
Numerous research has been conducted to implement privacy preservation in fields involving privacy-sensitive data. These endeavors primarily strive to develop privacy-preserving technologies, which can be broadly classified into general techniques and domain-specific techniques.

\subsubsection{General Technique}
Newton et al. \cite{ppdiface2005newton} proposed k-Same, which preserves sufficient facial details while preventing faces to be recognized, thus protecting personal privacy in video surveillance. Zhou et al. \cite{passbio2018zhou} launched a user-centric model PassBio based on the Threshold Predicate Encryption (TPE) scheme. This model enables users to encrypt individual biometric templates, also avoid exposing the plaintext in similarity computing during online authentication. Wu et al. \cite{attacks2023wu} discovered two impersonation attacks that could easily cheat the PassBio. They then presented a Verifiable Threshold Predicate Encryption (VTPE) scheme to effectively detect these attacks with higher security strength.

\subsubsection{Domain-Specific Technique}
This kind of schemes are usually application-oriented in specific domains. In medical diagnosis, researchers employed federated learning in data sharing and collaboration \cite{kaissis2020secure,bai2021advancing}, image reconstruction for patient facial image deidentifying \cite{yang2022digital}, image retrieval for encrypted medical image search \cite{ppir2019shen}, \emph{etc}. In IoT and blockchain, there are studies including privacy protection in smart systems \cite{kosba2016blockchain,ppsmartgrid2019gai}, encrypted IoT data sharing based on blockchain \cite{iotoverview2020huo,lu2020blockchain}, \emph{etc}. In cloud applications, multi-keyword search over encrypted cloud data \cite{clouddata2014cao} and content-based cloud image retrieval schemes \cite{cloudcompute2016xia} are explored under privacy-preserving scenarios. In social network service, anonymized social networks \cite{negihborattack2008zhou}, customized network data utility \cite{customized2018he}, and network personalization \cite{wu2022federated} have been studied to guarantee privacy security.

Nevertheless, no research has explicitly studied the privacy violation issue in online handwriting verification. Even though existing online biometric authentication schemes \cite{passbio2018zhou,outsourced2018hu,attacks2023wu} can be extended to online handwriting verification, their proposed methods may not adequately cope with all potential threats of privacy breaches, such as more sophisticated malicious privacy attacks. Instead of proposing dedicated schemes to balance privacy with the need of data utilization, we attempt to safeguard personal private information at its root, \emph{i.e.}, leveraging handwriting data with innate privacy-preserving nature. The proposed online RDS affords an inherently appropriate and straightforward solution for this issue, manifesting huge potential in applications.

\subsection{Online Character/Digit/Digit String Dataset}
Recently, researchers have gradually shifted attention toward exploring new biometric traits for identity verification, rather than limiting to the conventional signature. Nguyen et al. \cite{nguyen2017draw} collected a non-public digit dataset to evaluate Personal Identification Number (PIN) authentication. Tolosana et al. \cite{tolosana2019biotouchpass} proposed the e-BioDigit database, which comprises 7,440 online handwritten digits written by finger on mobile equipment. They required users to finger-draw generated One Time Passwords (OTP) in their touch biometric system as a second level of authentication for higher security. In addition, they published the MobileTouchDB database in \cite{tolosana2020biotouchpass2}, where users were required to draw digits and characters and repeat a 4-number password (always ``5 7 8 4'') in different ways. Zhang et al. \cite{zhang2022msds} created the MSDS dataset, whose MSDS-TDS subset novelly utilized Token Digit Strings for handwriting verification, with 402 users providing 20 genuine/forged samples in two captured sessions. This is the first study investigating continuous-handwritten digit string in handwriting verification. Besides, there are several digit datasets that were not primarily intended for identity verification, but could be potentially used for authentication after proper re-curation. El et al. \cite{el2008off} presented an Arabic handwriting dataset LMCA, including 30,000 online modality digits collected from 55 writers. Azeem et al. \cite{arabicdigit2012azeem} proposed an online Arabic handwritten digit dataset AOD for recognition, containing 30,000 samples from 300 writers. Njah et al. \cite{multidigit2012njah} collected a multi-language dataset MAYASTROUN, including 6,500 digits written by 355 writers.

Conceptually, the OTP in \cite{tolosana2019biotouchpass} seems similar to our proposed RDS as the content is random. However, the verification system can merely analyzes isolated digits in OTP rather than an integral digit sequence. Although one can generate integral digit strings by sequentially stitching digits, the spliced ones can not substantially substitute consecutively written RDS as they lack natural transitions and connections. Therefore, we opt for coherent-written RDS to more accurately capture inherent personal writing features. In addition, the aforementioned digit datasets lack corresponding skilled forgery data, thus are not available in handwriting verification as anti-skilled forgery evaluations are required.

\subsection{Online Handwriting Verification}
Online handwriting verification has seen rapid advancement in recent decades, especially online signature verification \cite{diaz2019aperspective}. Mainstream methods can be grouped as traditional statistical feature extraction-based methods and deep learning-based methods, in which most of the former are dominated by Dynamic Time Warping (DTW) \cite{vintsyuk1968speech}, while the latter typically harness CNN/RNN for local/global feature representation learning and achieve state-of-the-art.

\subsubsection{Statistical Feature Extraction-Based Method}
Conventional approaches usually involve three steps: first, statistical/hand-crafted features extraction; second (optional), feature encoding; third, decision making. The statistical features include time function features, such as coordinate, velocity, sine/cosine measures, and their high-order derivatives \cite{symbolic2009guru,gmmdtw2017sharma,gmm2017yang,tang2018info,hmm2018farimani,okawa2021time}; frequency features, such as the discrete Fourier/Cosine/Wavelet/Hartley/Walsh-Hadamard/Kreke transformation features \cite{dctdwtdft2018miaba,gmm2017yang,hmmdwt2017chavan}; principal features decomposed by Singular Value Decomposition (SVD) \cite{svd2008kamel}; Histogram of Oriented Gradient (HOG) \cite{hog2019mshengu}, \emph{etc}. For the second step, some studies employed the Gaussian Mixture Model (GMM) \cite{gmmdtw2017sharma,gmm2017yang,gmm2018sharma} to encode the pre-extracted statistical features. Notably, this step is not always required. Mainstream decision-making schemes include the Support Vector Machine (SVM) \cite{svm2011kour,svm2012parodi}, Hidden Markov Model (HMM) \cite{fierrez2007hmm,maiorana2008tem,hmmdwt2017chavan,hmm2018farimani}, and DTW \cite{dtw1990parizeau,dtw2011yu,kolmogorovdistance2014griechisch,tang2018info,twed2020auddya,okawa2021time}. DTW exhibits exceptional non-linear modeling capability in sequence comparison and is therefore popularly adopted to measure the similarity between the query and template in online signature verification. Beyond utilizing the vanilla DTW, much research effort has been devoted to improving the performance of this algorithm. Yu et al. \cite{dtw2011yu}, Griechisch et al. \cite{kolmogorovdistance2014griechisch}, and Auddya et al. \cite{twed2020auddya} respectively proposed to utilize the Mahalanobis distance, Kolmogorov-Smirnov distance, and Time Warp Edit distance as the inner distance function in DTW for better online trace similarity calculation. Tange et al. \cite{tang2018info} developed a computation-efficient FBU-DTW with a simplified backward distance calculation process. Okawa \cite{okawa2021time} proposed local stability weighted-DTW by incorporating the local stability sequence as weights for vanilla DTW, where the local stability is the averaged direct matching points between the mean template set and all reference signatures.

\begin{figure*}[t]
	\centering
	\includegraphics[width=0.92\linewidth]{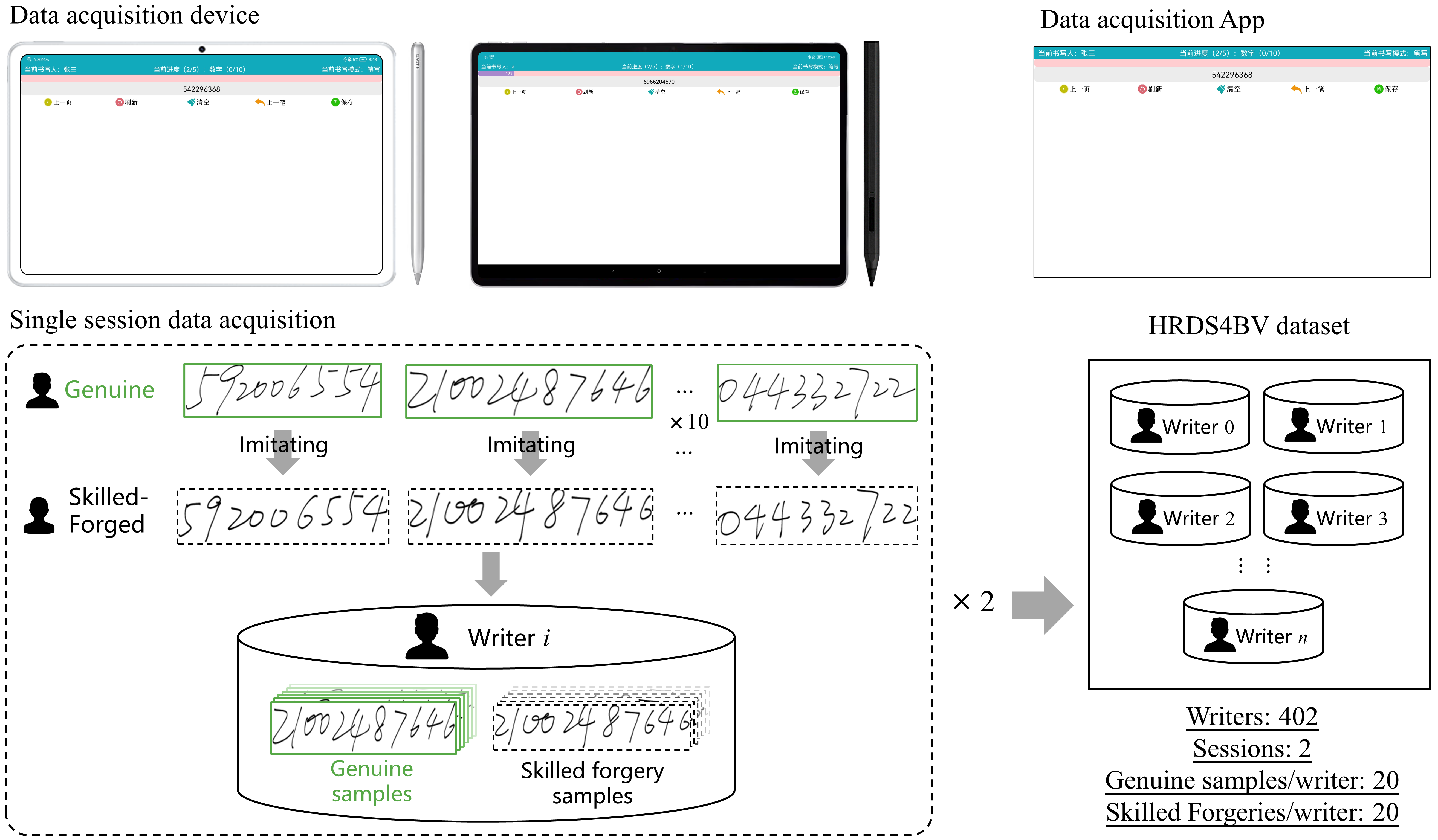}{}
	\caption{Illustration of the construction details of the HRDS4BV dataset, including data collection devices, the graphical user interface of the data acquisition app, and the acquisition process.}
	\label{Fig::acquisition}
\end{figure*}

\subsubsection{Deep Learning-Based Method}
The prevalence of deep learning has reinvigorated this field with new and efficient modeling schemes. Models based on CNN/RNN have exhibited outstanding performances in this task and recently dominate the state-of-the-art. Deep model-based approaches can be grouped into two types. The first type targets local feature modeling, which usually combines deep models with DTW \cite{wu2019DeepDTW,wu2019prewarp,tolosana2021deepsign,jiang2022dsdtw}. Wu et al. \cite{wu2019prewarp} and Tolosana et al. \cite{tolosana2021deepsign} used DTW to pre-align time sequences before inputting them to siamese CNN/RNN-based networks. Wu et al. \cite{wu2019DeepDTW} adopted DTW on top of a Siamese CNN to align the two inputs before loss optimization. Jiang et al. \cite{jiang2022dsdtw} introduced soft DTW, a differentiable formulation of vanilla DTW, and packed it in the loss function to achieve better end-to-end optimization. The second type learns global feature representations \cite{lai2018recurrent,park2019robust,ahrabian2019usage,vorugunti2019osv,li2019stroke,lai2021synsig2vec}, which mostly adopted RNN. Lai and Jin \cite{lai2018recurrent} proposed a gated auto regressive unit and a Recurrent Adaption Network for global representation learning. They also proposed a new length-normalized path signature descriptor to better holistically describe online signatures. Li et al. \cite{li2019stroke} first modeled divided stroke patches then the entire signature using RNN. Similarly, Park et al. \cite{park2019robust} learned representations at both stroke and signature levels, while using a LSTM-CNN network. Sekhar et al. \cite{vorugunti2019osv} manually extracted global features, \emph{e.g.}, max velocity, average pressure, and input them to a CNN model. Lai et al. \cite{lai2021synsig2vec} proposed a CNN-based model Sig2Vec and a selective pooling module to pool the feature sequence with multi-head attention \cite{attention2017vaswani} and learnable spatial weights. It aggregates all local subspace features and forms a global feature vector with the awareness of the context.

Zhang et al. \cite{zhang2022msds} have proved that online signature verification methods can be adapted to other handwriting verification tasks, such as online digit string verification. However, these methods heavily depend on fixed content and neglect style modeling, leading to huge accuracy decline when transferred to RDS verification due to the absence of content information. This discrepancy in model capability between text-dependent and text-independent scenarios emphasizes the necessity for specialized approaches. Accordingly, we propose the PAVENet model and the DPM module dedicated to enhance handwriting pattern learning and style modeling in text-independent writing scenarios.

\begin{figure*}[t]
	\centering
	\includegraphics[width=0.97\textwidth]{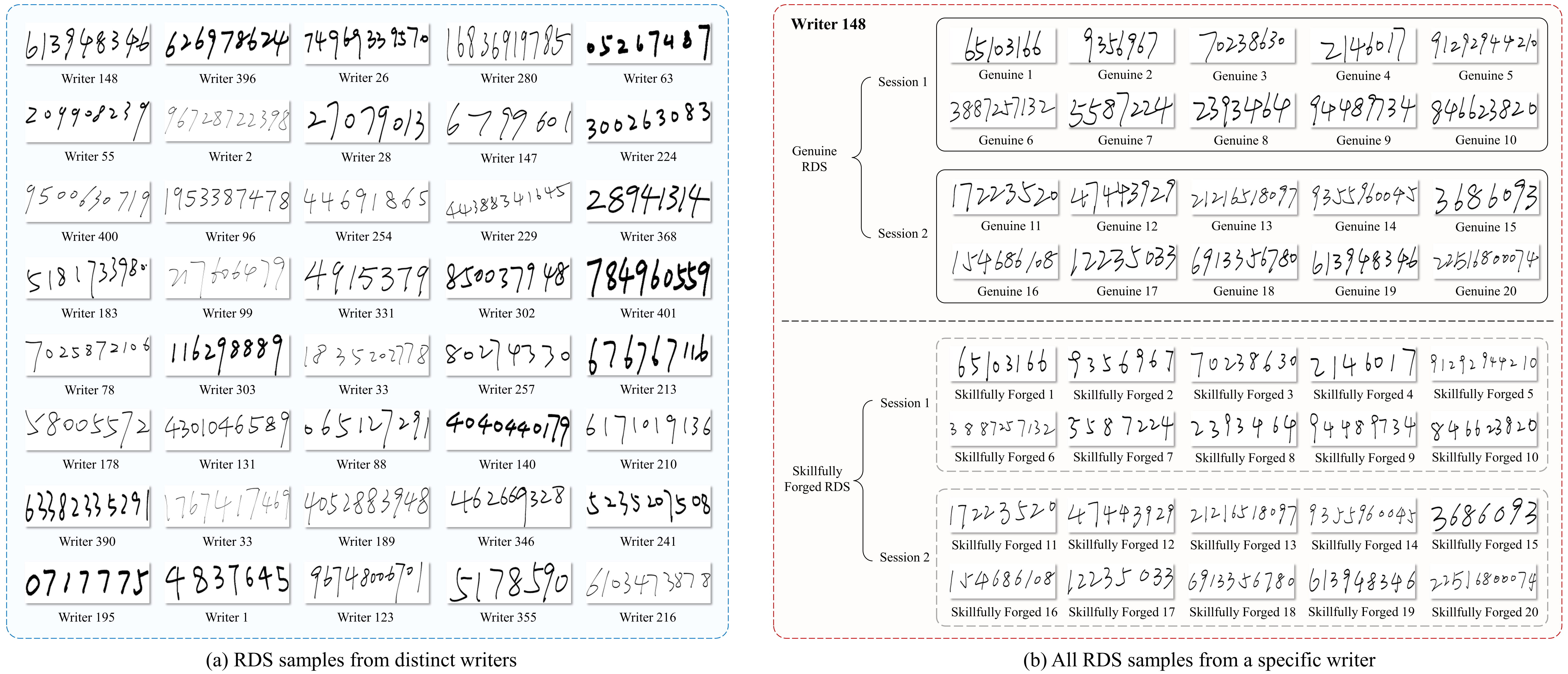}{}
	\caption{Data visualization. (a) Visualization of data diversity. The samples are randomly selected from distinct writers, showcasing obvious inter-writer variations in writing styles. (b) Visualization of data representativeness. We exhibit all genuine samples and skilled forgeries from a random writer. The contents and lengths of genuine samples vary since they are randomly assigned. Each skilled forgery holds a one-to-one correspondence to a specific genuine sample. The data were collected in two sessions separated by at least three weeks, with 10 genuine and 10 skillfully forged samples per session.}
	\label{Fig::data_example}
\end{figure*}

\begin{figure*}[t]
	\centering
	\includegraphics[width=0.95\textwidth]{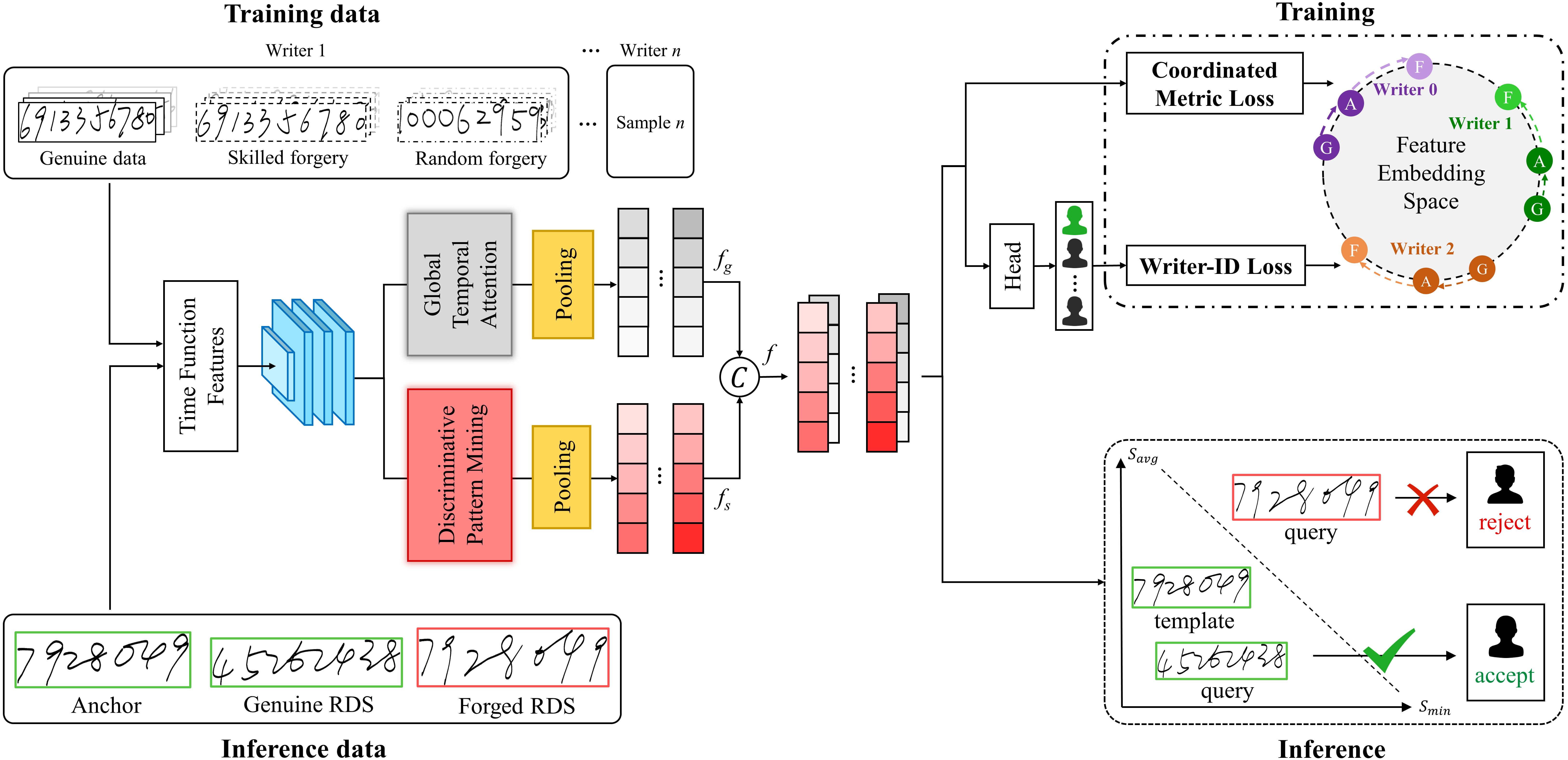}{}
	\caption{Overall framework of the proposed PAVENet. Top: model training. The training data includes genuine RDS, the corresponding skilled forgeries per writer, and random forgeries. The model is trained to distinguish genuine and forged RDS through feature space optimization. Middle: a simplified model architecture of the PAVENet. The PAVENet is dedicatedly designed to capture discriminative handwriting patterns, bolstering the learning of unique writing styles. Bottom: model inference. The model compares the query RDS against the template to determine whether to accept or reject the query.}
	\label{Fig::overall}
\end{figure*}

\section{Privacy Threat}
\label{sec::threat}
Handwriting verification systems confront various privacy concerns, with the most prominent being the inclusion of personal information within the handwriting content. We formally define several key privacy threat models as follows. (1) \textbf{Personally identifiable information leakage.} Traditional handwriting verification (\emph{e.g.}, signature/Token Digit String (TDS)-based) risks exposing sensitive personal data embedded in the content, such as names or phone numbers \cite{lai2021synsig2vec,deepwritesyn2021aaai,zhang2022msds}. (2) \textbf{Replay attacks.} Adversaries might reuse the previous RDS sample to gain unauthorized access \cite{zhang2015strong}. (3) \textbf{Forgery (adversarial) attacks.} Typical forgery attacks include skilled and random forgeries \cite{diaz2019aperspective,attacksurvey2021impedovo}. Skilled forgeries are impersonated handwriting that replicates genuine users' writing styles based on known samples. Random forgeries are authentic samples from writers other than the stated one.

To mitigate these issues, we propose using Random Digit Strings (RDS) for handwriting verification. RDS consists of arbitrary digits, which renders it text-independent and devoid of personally identifiable content. This biometric inherently prevents personal information leakage and achieves privacy preservation. Also, by requiring users to sign a new RDS with different content each time, the system naturally thwarts replay attacks, a vulnerability persisting in conventional signature/TDS verification due to the static nature of their content. Additionally, we design a dedicated Pattern Attentive VErification Network (PAVENet) to bolster the capture of consistent and salient style writing features in RDS. This approach effectively increases the resilience against both types of forgery attacks, building a reliable RDS verification system. The details of RDS data collection and the development of PAVENet are elaborated in the subsequent Sec.~\ref{sec::create_dataset} and Sec.~\ref{sec::method}, respectively.

\section{HRDS4BV Dataset}
\label{sec::create_dataset}
In this section, we describe the construction of the HRDS4BV dataset in detail. An intuitive illustration of the acquisition process is depicted in Fig.~\ref{Fig::acquisition}. To ensure data diversity and representativeness, we have considered the following key factors during dataset construction.

First, we emphasize the naturalness of online handwritten RDS, \emph{i.e.}, utilizing naturally and coherently handwritten RDS rather than the synthesized counterparts, to maximumly preserve intrinsic writing characteristics. While it is possible to synthesize visually resembling RDS by spatially splicing separate digits, this approach treats each digit independently, without considering the context or neighboring parts. Consequently, such synthesized RDS may lack smooth transitions and contextual connections between individual digits, resulting in the loss of crucial features associated with stroke flow and pen movement, thereby undermining the analysis of latent writing characteristics. In contrast, consecutively written RDS maintains the natural flow and contextual continuity of handwriting, innately showcasing natural variability and idiosyncrasies of the writer's unique writing characteristics. Hence, naturally handwritten RDS provides more comprehensive features and would fundamentally bolster the verification system built based upon it. Therefore, we prioritize the utilization of natural RDS to capture the essence of an individual's handwriting, enabling more accurate and reliable verification.

Second, we meticulously consider the content characteristics of RDS in the dataset. To guarantee its privacy-preserving property, each RDS is generated by combining digits from 0 to 9 in a random order. Therefore, every RDS has completely different content. To increase diversity, we vary the lengths of RDS from 7 to 11 digits. In addition, we consider the inter-session variability problem, which pertains to the inherent natural fluctuations and dissimilarities observed in an individual's handwriting across different writing sessions, stemming from a range of circumstances or psychological factors. Previous studies have proved that this variability is non-negligible for behavioral biometrics \cite{kholmatov2009susig,tolosana2019biotouchpass,zhang2022msds}, and is also a common occurrence in practical applications. Hence, we set up two acquisition sessions with a time interval of over 21 days to account for this aspect. An equal number of samples were acquired per writer per session to ensure consistency. By incorporating this setting, the HRDS4BV dataset aligns with real-world application requirements and facilitates more comprehensive evaluations regarding this critical challenge.

Third, we carefully design the data collection process to ensure data quality. For online data acquisition devices, we chose two types of Android tablets equipped with specific styluses. Concurrently, we developed a custom handwriting collection app to serve as the dynamic signal recording system. The tablets, styluses, and app cooperatively simulate an immersive writing process akin to real-world scenarios for the participants, preserving the naturalness of handwriting to the fullest extent possible. Writers were instructed to handwrite the RDS using styluses within a designated area in the app, according to the predefined content displayed by the app. A new RDS will be updated after the former one has been saved. The app automatically logs the dynamic information produced in the writing process, including $x$ coordinates, $y$ coordinates, pressures, timestamps, pen-ups, and pen-downs.

We have collected handwritten RDS from 402 writers to form the HRDS4BV dataset. In each acquisition session, every writer has produced 10 genuine samples and 10 skilled-forged samples. To obtain skilled forgeries, each writer was tasked with imitating all 10 samples of another randomly selected writer. The imitator was required to reproduce the same-content RDS and emulate the original writing style through watching the screen recording of the original writing process and repeatedly practicing. To ensure uniqueness, a strict one-to-one correspondence was maintained between the imitator and the imitated, with each imitator assigned a specific writer to mimic. As a result, after completing two capture sessions, each writer possesses a total of 20 genuine samples and 20 skilled forgeries in the dataset, culminating in the HRDS4BV dataset comprising 8,040 genuine samples and 8,040 corresponding skillfully forged counterparts. It is noteworthy that acquiring skilled forgery data is the most time-consuming and labor-intensive part of the entire data collection process. To uphold data quality, expert supervisors were present throughout the process to provide necessary guidance. Following the completion of data collection, all data underwent rigorous correction and cleaning procedures to ensure its correctness and high quality. 

In Fig.~\ref{Fig::data_example}, we present the visualization of partial samples from the HRDS4BV dataset. Fig.~\ref{Fig::data_example} (a) displays randomly selected samples from 40 distinct writers, vividly illustrating the diversity in writing styles within the dataset. Fig.~\ref{Fig::data_example} (b) exhibits all RDS samples from an exemplar writer, providing an intuitive manifestation of the data properties and representativeness of our dataset. Additionally, we summarize the details of the HRDS4BV dataset and perform comparisons with existing digit/digit string datasets in Table~\ref{Table::comparison}.

\section{Methodology}
\label{sec::method}
\subsection{Overall Framework}
\label{sec::overall}
We present the architecture of the proposed Pattern Attentive VErification Network (PAVENet) in Fig.~\ref{Fig::overall}, including both training and inference procedures. The PAVENet takes preprocessed online RDS as input and outputs representative feature vectors. To better reflect real-world scenarios, we frame RDS verification as an open-set task, where the model is required to identify unseen writers outside the training data. Thus we adopt the metric-learning optimization strategy for model training. During inference, we adaptively compute a threshold to divide the genuine and forged samples. A query sample whose distance to the template falls below the threshold will be accepted as a genuine sample, otherwise, it is regarded as a forgery and rejected.

\subsection{Backbone Network}
\begin{figure}[t]
	\centering
	\includegraphics[width=\linewidth]{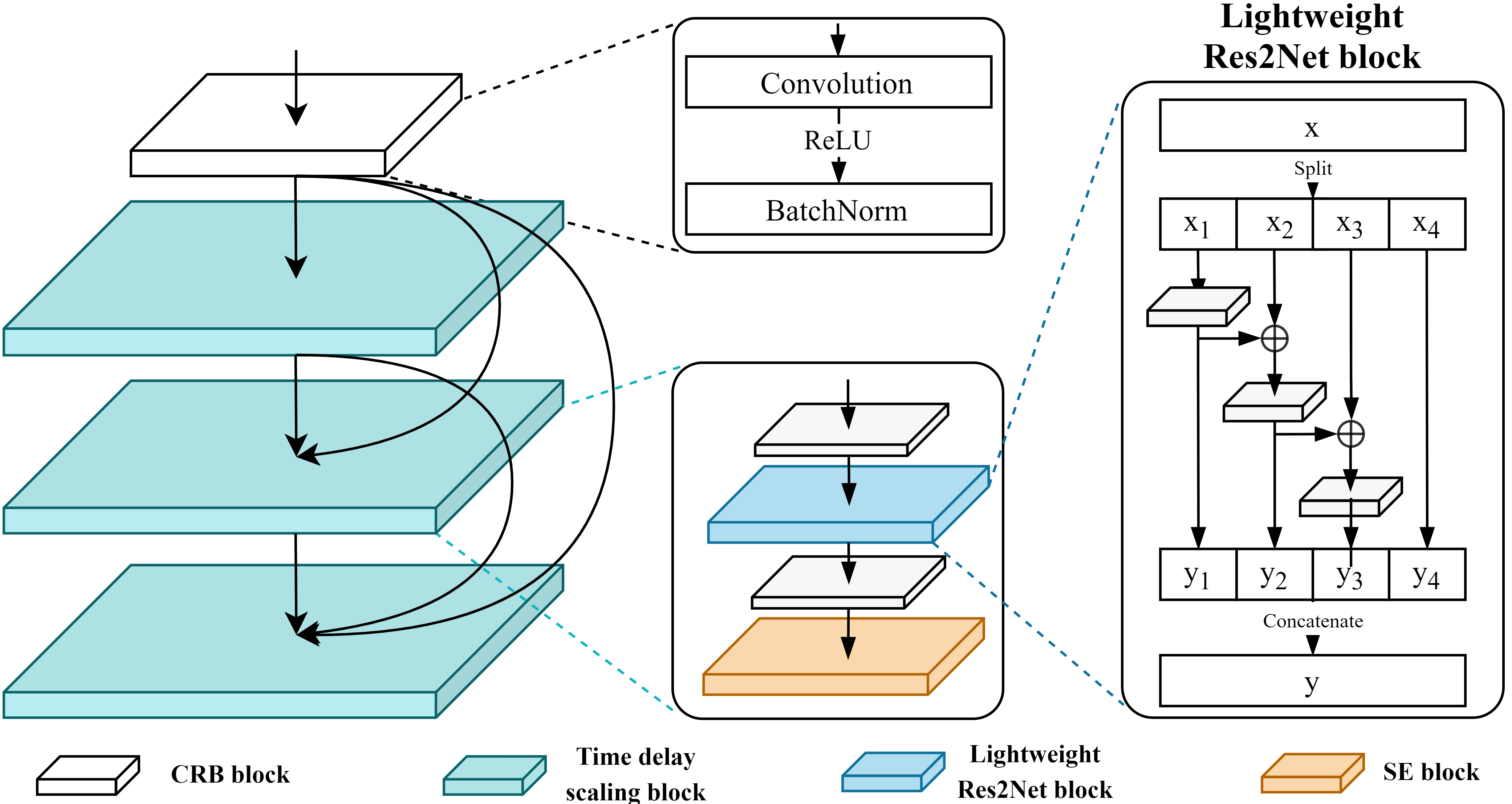}{}
	\caption{Architecture of the backbone network.}
	\label{Fig::arch}
\end{figure}

Fig.~\ref{Fig::arch} illustrates the backbone architecture of the PAVENet. The backbone is inspired by Res2Net \cite{gao2021res2net}, DenseNet \cite{huang2017densely}, and ECAPA-TDNN \cite{desplanques2020ecapa}. Res2Net is a well-established and powerful model in vision tasks benefiting from the most pivotal two-dimensional Res2Net block. We adapt and customize the corresponding one-dimensional Res2Net block to suit sequence modeling, and further introduce the light-weight Res2Net block, a lighter version of the vanilla one, in which the original $ 1 \times 1$ convolutions at the front and the end are removed to reduce computational overheads. Further, we concatenate one \emph{CRB} block and three time delay scaling blocks by dense connection \cite{huang2017densely} for knowledge reuse. Given a batch of RDS $x \in \mathbb{R}^{12 \times L}$ (12 is the channel number, corresponding to 12 time function features that we will illustrate later in Section~\ref{sec::preprocess}; $L$ is the maximum length of this batch) as input, we first project it to the fixed 512 channels through a \emph{CRB} block. Subsequently, $x$ is split to $x_i, i \in \{1,2,3,4\}$ and each of the first three feature subsets $x_i, i \in \{1,2,3\}$ are processed by a \emph{CRB} block. The output is termed $y_i,i \in \{1,2,3,4\}$, which are then concatenated as $y$. The output $y$ is sent to the following modules for further feature modeling.

\subsection{Discriminative Pattern Mining}
\begin{figure}[t]
	\centering
	\includegraphics[width=\linewidth]{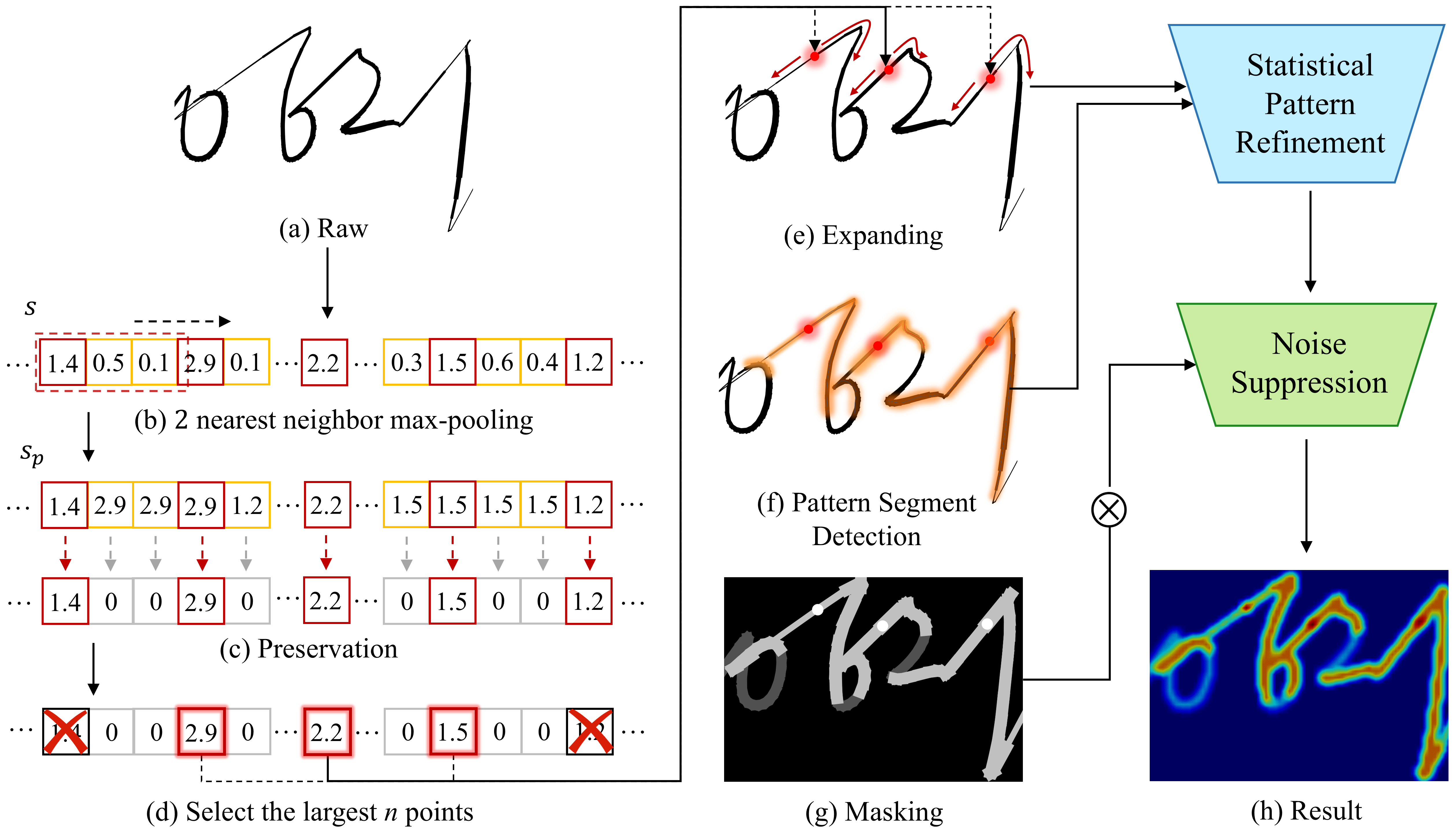}{}
	\caption{Schematic of the Discriminative Pattern Mining (DPM) module.}
	\label{Fig::dpm}
\end{figure}

Since the high variation of unconstrained-written RDS poses great challenges in identifying personal writing styles, prevailing methods suffer from significant performance degradation owing to the neglect and deficiency in style modeling. We therefore propose a new Discriminative Pattern Mining (DPM) module to explicitly excavate handwriting patterns and empower style representation. The schematic of DPM is illustrated in Fig.~\ref{Fig::dpm}. Given an input feature sequence $s \in \mathbb{R}^{512 \times L}$ (512 is the channel number and $L$ is the maximum length of this batch), we begin by extracting pattern key points via max-pooling:
\begin{equation}
	s_p = \{max(s_{[j:j + K)})\}, j \in \{0,1,...,L-K+2\},
\end{equation}
\noindent where $K$ is the kernel size for max-pooling. From $s_p$, we retain the time steps that originate from $s$ and select the largest $n$ points as pattern key points of current handwriting, denoted as $p_i, i \in \{1,2,...,n\}$. The pooling and selection steps are illustrated in Fig.~\ref{Fig::dpm} (b), (c), and (d).

Subsequently, we extract the pattern segments $ps_i \in \mathbb{R}^{512 \times l_s}, i \in \{1,2,...,n\}$ as depicted in Fig.~\ref{Fig::dpm} (e) and (f). Each segment is centered on a corresponding key point $p_i$ and extends on both sides symmetrically, with a length of $l_s = \lfloor \lfloor L / 4 \rfloor / n \rfloor$. These segments capture local writing patterns around each key point, encapsulating the unique characteristics of the writer's handwriting style.

To further emphasize the discriminative writing patterns, we calculate the mean $\mu$ and standard deviation $\sigma$ of all segments, using them to perform statistical pattern refinement on the holistic handwriting sequence $s$:
\begin{equation}
	\mu = \frac{1}{N} \sum_{i,j,k} ps_{i,(j,k)},
\end{equation}
\begin{equation}
	\sigma = \sqrt{\frac{1}{N} \sum_{i,j,k} (ps_{i,(j,k)} - \mu)^2},
\end{equation}
\begin{equation}
	\widetilde{s} = (s - \mu) / \sigma,
\end{equation}
\noindent where $N=512 \times l_s \times n$, $j,k \in \mathcal{S}_{512,l_s}$. With the refined sequence, we create a positional mask $m$ (depicted in Fig.~\ref{Fig::dpm} (g)) to focus on the key segments and suppress noise by assigning higher importance to the segment part and lowering the influence of irrelevant parts:
\begin{equation}
	\label{Eq::mask}
	m(t) = 
	\begin{cases}
		0.9 & u_i \leq t \le v_i\\
		0.005 & \text{otherwise},
	\end{cases}
\end{equation}
\begin{equation}
	\label{Eq::suppress}
	\widetilde{s} = \widetilde{s} \odot m.
\end{equation}

Finally, the refined $\widetilde{s}$ is combined with the original $s$ through residual connections to get a new $\widetilde{s}$. We use a selective pooling \cite{lai2021synsig2vec} module to transform $\widetilde{s}$ into fixed-length vectors $f_s$ that represent the style features.

Intuitively, due to the long-established writing habits, handwriting patterns of the same individual are well-formed and consistent. From the perspective of human visual perception, when observing someone's handwriting, we naturally notice these conspicuous and distinctive writing patterns, such as stroke hyphenations and stroke flourishes. They usually encapsulate rich writing characteristics and could serve as valuable representations of personal writing style. To effectively capture these distinctive features, we leverage max-pooling for its exceptional ability to accentuate informative regions \cite{woo2018cbam,zhou2019objects}. In this case, the pattern key points and pattern segments are adaptively computed according to distinct and proprietary personal writing characteristics from a high-dimensional perspective, reflecting the discriminative styles of personal handwriting. We further apply the statistical pattern refinement to strengthen the detected pattern features. After the pattern refinement, points in $s$ but not in $ps$ would be contaminated and become noise. So we generate $m$ and multiply them with a small factor of 0.005 to filter out the noise from these points. $K$ is set to 3 and $n$ is set to 8.

\subsection{Gloabl Temporal Attention}
Inspired by \cite{letter2021chen}, we design a Global Temporal Attention (GTA) module to model the long-term dependency of dynamic handwriting. Denote the input feature sequence as $y$, we use a long-range dependency extractor $\phi$ for global modeling and compute the relative attention weights as:
\begin{equation}
	\label{Eq.1}
	\hat{y} = \phi(y),
\end{equation}
\begin{equation}
	\label{Eq.2}
	attn = softmax(\hat{y}),
\end{equation}
\noindent Among the four candidates: GRU, LSTM, Temporal Convolutional Neural Network layer \cite{tcn2018bai}, and standard Transformer layer \cite{attention2017vaswani}, we choose LSTM as $\phi$ for its superior performance. The $attn$ represents the relative importance of each time step within the entire RDS. Then we perform element-wise multiplication of $y$ and $attn$, followed by a residual connection to obtain $\widetilde{y}$:
\begin{equation}
	\label{Eq.3}
	\widetilde{y} = y \odot attn,
\end{equation}
\begin{equation}
	\label{Eq.4}
	\widetilde{y} = \widetilde{y} + y.
\end{equation}
$\widetilde{y}$ is then input to a selective pooling module to get fixed-length vectors $f_g$ that represent the global features. $f_s$ and $f_g$ are then concatenated as $f$ (Fig.~\ref{Fig::overall}) for loss optimization.

\subsection{Model Optimization}
To approach real-world applications, the PAVENet is hoped to develop the ability to verify previously unseen writers. We therefore train the model in an open-set manner, in which writers of the training and test set are mutually exclusive. First, we adopt the metric learning-based optimization scheme, amalgamating the lifted structure triplet loss \cite{song2016deep} $\mathcal{L}_{tri}$ and the N-pair\&angular loss \cite{wang2017angular} $\mathcal{L}_{N\&ang}$ to form the coordinated metric loss $\mathcal{L}_{CM}$
\begin{equation}
	\mathcal{L}_{CM} = 0.1\mathcal{L}_{tri} + \mathcal{L}_{N\&ang}.
\end{equation}
\noindent The coefficient $\lambda$ in $\mathcal{L}_{N\&ang}$ is set to 1.

Subsequently, to prevent random forgery attacks, the model needs to classify the writer's identity (writer-ID) to who the RDS belongs. We adopt the standard cross entropy loss as the writer-ID Loss. The \emph{Head} module (Fig.~\ref{Fig::overall}) is a multi-layer perceptron and is utilized to transform the concatenated feature vectors $f$ to probabilities $p$ for loss computation:
\begin{equation}
	\mathcal{L}_{ID} = -\frac{1}{N_c} \sum_i^{N_c} p_i log(q_i),
\end{equation}
\noindent where $q_i$ denotes the distribution of ground truth and $N_c$ denotes the number of all writer identities.

Finally, the full optimization objective is:
\begin{equation}
	\label{Eq.23}
	\mathcal{L} = \mathcal{L}_{CM} + \mathcal{L}_{ID}.
\end{equation}

\section{Experiment}
\label{sec::experiment}
\subsection{Dataset}
\label{sec::exp_dataset}
We conduct experiments on the HRDS4BV dataset and the e-BioDigit database \cite{tolosana2019biotouchpass} to evaluate the effectiveness of RDS and the proposed PAVENet. To investigate the naturalness of handwritten RDS and the effect of RDS's length, we respectively generate two synthetic experimental datasets, named e-BioDigit-L and e-BioDigit-S. These datasets are created by horizontally splicing single digits of the e-BioDigit database and possess different lengths of RDS. Each user of the e-BioDigit has eight finger-written samples from 0-9, in which the former and the latter four were captured in two separate sessions. We respectively synthesize 10 strings using digits of each session to match the data volume of the HRDS4BV dataset. Since e-BioDigit only provides $x$ and $y$ coordinates but no pressure information, we simply add a subsequent pressure dimension and set the value as 1. Note that no skilled forgery data can be synthesized as there are no forged digits in the e-BioDigit. The e-BioDigit-L dataset consists of strings with lengths ranging from 7 to 11 (the same length interval as HRDS4BV), and the e-BioDigit-S dataset is comprised of strings with lengths between 3 to 4.

\subsection{Protocol}
\label{sec::protocol}
To perform comprehensive evaluations, we design the following experiment protocols:
\begin{itemize}
	\item \textbf{Data splitting.} The data splitting is performed to approximate real-world application scenarios, with a near 1:1 ratio for training and testing data. For the HRDS4BV dataset, we follow the setting of \cite{zhang2022msds}, designating the data of the first 202 writers for training and the remaining 200 writers for testing. This results in 8,080 training samples and 8,000 testing samples. For e-BioDigit-L and e-BioDigit-S, we adhere to the settings established by the original paper \cite{tolosana2019biotouchpass}, dividing the first 50 writers for training and the rest 43 writers for testing. This results in 1,000 training samples and 860 testing samples for each dataset.
	\item \textbf{Metrics.} We adopt the EER\% as the evaluation metric. The Equal Error Rate (EER\%) refers to the point where the False Rejection Rate (FRR\%) is equal to the False Acceptance Rate (FAR\%), which is a standard and commonly used metric for handwriting verification \cite{diaz2019aperspective,lai2017online,lai2018recurrent,wu2019DeepDTW,wu2019prewarp,lai2021synsig2vec,tolosana2021deepsign,jiang2022dsdtw,zhang2022msds}. The experimental results are all displayed in the format of $EER_{g}/EER_{l}$, in which the former is computed under a global threshold and the latter is computed under a local (user-specific) threshold \cite{lai2021synsig2vec,zhang2022msds}.
	\item \textbf{Imposter types.} We consider both skilled and random forgeries as impostor types, as outlined in Sec.~\ref{sec::threat}. Skilled forgeries are provided in the HRDS4BV dataset while random forgeries are selected from authentic samples of writers other than the current one.
	\item \textbf{Inference and verifier.} To compute the EERs, we use the template- and distance-based verifier proposed by Lai et al. \cite{lai2021synsig2vec}. During the verification phase, the verifier computes the distance between the enrolled template and the input query \cite{template2013sae,tang2018info,okawa2021time,lai2021synsig2vec}. The query sample will be accepted if the distance falls below a pre-determined threshold, as illustrated in the right bottom panel of Fig.~\ref{Fig::overall}. We discuss the template selection strategy in the subsequent part. In experiments, all models output final feature vectors, which are then input to the verifier to obtain EERs.
	\item \textbf{Template selection.} As previously reported \cite{tolosana2019biotouchpass,tolosana2021deepsign,lai2021synsig2vec,jiang2022dsdtw,zhang2022msds}, the number of templates used for verification can significantly influence the performance of EER. Following most existing methods \cite{tolosana2021deepsign,lai2021synsig2vec,jiang2022dsdtw}, in our experiments, we use either four templates or a single template against one query for testing, termed as 4 vs 1 or 1 vs 1, respectively. When testing with across-session RDS in the 4 vs 1 case, the four templates are the first two samples of each session.
\end{itemize}

\subsection{Data Preprocessing}
\label{sec::preprocess}
\begin{table}[h]
	\renewcommand{\arraystretch}{1.2}
	\caption{Time function features.}
	\label{Table::timefunc}
	\centering
	\begin{tabular}{c c}
		\toprule
		\noalign{\vspace{-2pt}}
		\# & Features\\
		\hline
		1 & First-order derivative of Coordinate $x$: $\dot{x}$\\
		\hline
		2 & First-order derivative of Coordinate $y$: $\dot{y}$\\
		\hline
		3 & Velocity magnitude: $v = \sqrt{\dot{x}^2 + \dot{y}^2}$\\
		\hline
		4 & Path-tangent angle: $\theta = \arctan{\frac{\dot{y}}{\dot{x}}}$\\
		\hline
		5 & Cosine of the path-tangent angle: $\cos{\theta}$\\
		\hline
		6 & Sine of the path-tangent angle: $\sin{\theta}$\\
		\hline
		7 & First-order derivative of $v$: $\dot{v}$\\
		\hline
		8 & First-order derivative of $\theta$: $\dot{\theta}$\\
		\hline
		9 & Log curvature radius: $\rho = \log{\frac{v}{\dot{\theta}}}$\\
		\hline
		10 & Centripetal acceleration magnitude: $\bigtriangleup v =  v \cdot \dot{\theta}$\\
		\hline
		11 & Total acceleration magnitude: $a = \sqrt{\dot{v}^2 + \bigtriangleup v^2}$\\
		\hline
		12 & Pressure: $p$\\
		\noalign{\vspace{-2pt}}
		\bottomrule
	\end{tabular}
\end{table}
\unskip

We utilize the $x, y$ coordinates and pressure of the raw online RDS data as the initial features for further preprocessing steps. First, we normalize the handwriting coordinates. Given that each handwritten RDS is written on varying areas of the writing board with distinct sizes, the variation in size and location is unfavorable for learning spatial-invariant writing features. Hence we perform center normalization on $x$ and $y$ positions, shifting the center of the handwriting to (0,0), followed by scaling the $x,y$ coordinates to the range of (-1,1) while preserving their aspect ratio. Second, we apply min-max normalization on the raw pressure information. Then, we conduct a bi-cubic interpolation to double all the points. Further, we extract 12 time function features as detailed in Table~\ref{Table::timefunc} based on centered $x$, $y$, and normalized pressure. The time function features are eventually subjected to z-score normalization to have zero means and unit variance.

\subsection{Implementation Detail}
\label{sec::impl_details}
In each batch, we randomly sample data from two writers, with three genuine samples, three skilled forgeries, and three random forgeries for each writer, resulting in 2$\times$(3+3$\times$2)=18 samples in a single batch. The genuine samples and skilled forgeries are randomly selected from all genuine and forged samples of the current writer, respectively. The random forgeries comprise one randomly selected genuine sample from each of the other three writers excluding the current writer. Since RDS are of various lengths, they are end-padded with zeros to match the length of the longest sample within the batch. In addition, we employ a data augmentation technique, denoted as R-Del, which randomly deletes 5\%-7.5\% of the time points during training but not during testing. We train the model for 300 epochs and the initial learning rate is 0.001. We use the AdamW \cite{loshchilov2018decoupled} as the optimizer and the learning rate is decreased by multiplying a factor of 0.95 after each epoch.

\subsection{Investigating RDS for Identity Verification}
\subsubsection{Comparison with State-of-the-Art Methods}
\label{sec::hrds4bv}
\begin{figure}[t]
	\centering
	\includegraphics[width=\linewidth]{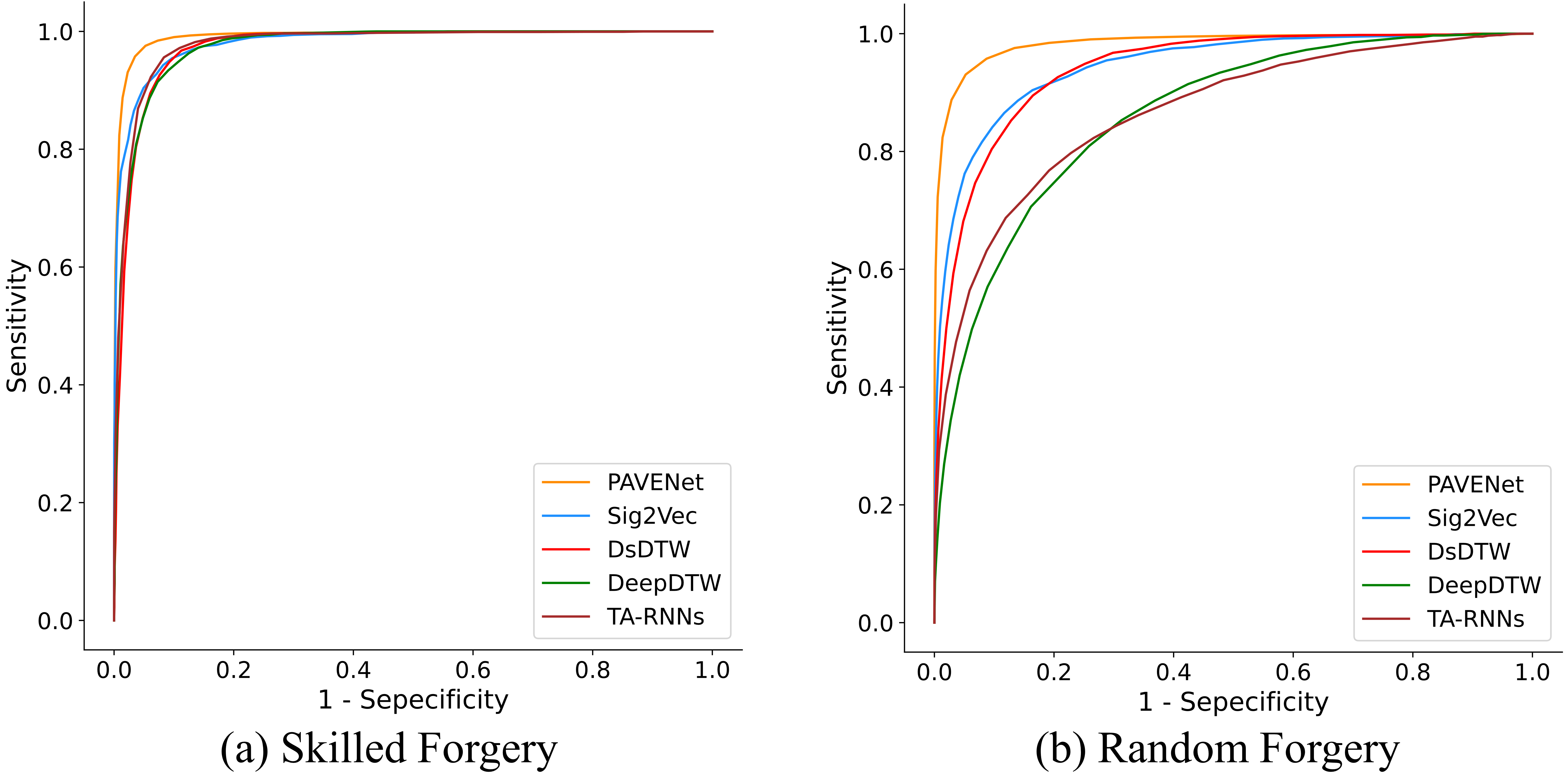}{}
	\caption{Receiver operating characteristic (ROC) curves for different methods tested on the across-session data of the HRDS4BV dataset under both skilled forgery and random forgery scenarios, using the 4 vs 1 template selection strategy.}
	\label{Fig::roc}
\end{figure}

We compare the performance of our proposed PAVENet against other state-of-the-art (SOTA) methods on RDS verification using the HRDS4BV dataset. DTW \cite{vintsyuk1968speech} denotes directly computing DTW distance on the input time function features for RDS matching without any training, whereas all other methods are trained deep models. The results are summarized in Table~\ref{Table::sota-hrds}. Session 1 or Session 2 indicates using solely the first/second session data for experiments, respectively, denoted as ``single-session'' in the following illustration. Session 1 \& 2 refers to employing data from both sessions collectively, denoted as ``across-session''. The data volume in single-session scenarios is half of that in across-session scenarios. Additionally, we plot the receiver operating characteristic (ROC) curves of different systems under skilled forgery and random forgery scenarios using across-session data in Fig.~\ref{Fig::roc}. As confirmed by the numerical results and ROC curves, the PAVENet yields optimal performances in both single-session and across-session evaluations. Specifically, on single-session data, PAVENet attains EERs of 1.86/0.39 ($EER_{g}/EER_{l}$) for skilled forgery and 2.77/1.26 for random forgery, outperforming the second-best performance of 3.15/0.80 and 6.27/5.01 by a substantial margin. Similarly, on across-session data, PAVENet achieves 3.92/2.33 and 6.24/5.15 for skilled forgery and random forgery, also significantly surpasses previous SOTA methods (6.77/4.93 and 13.30/11.77). These exceptional results affirm that despite greater variations and less available information in RDS, our proposed PAVENet consistently delivers promising results and remarkably outperforms preceding methods. We attribute the superior performances to PAVENet's capability to extract more comprehensive writing characteristics, especially the DPM module, which captures discriminative writing patterns and significantly enhances style modeling for unconstrained handwriting.

It is worth mentioning that, since most SOTA methods are built upon DTW \cite{wu2019DeepDTW,tolosana2021deepsign,jiang2022dsdtw}, their underperformance compared to PAVENet substantiates the inherent limitations of DTW-based methods to text-independent (TI) handwriting such as RDS. This inferiority is further exacerbated by the notably poor performance of vanilla DTW. Hence, it justifies the need for specialized methods like PAVENet to effectively address the challenging TI verification and calls for future efforts in this direction.

\begin{table}[t]
	\renewcommand{\arraystretch}{1.4}
	\caption{Comparisons with existing methods of verification Equal Error Rates (EER\%) tested on the HRDS4BV dataset. The results are displayed as $EER_{g}/EER_{l}$ and the optimal results are marked in \textbf{bold}. These conventions are maintained throughout the following results.}
	\label{Table::sota-hrds}
	\centering
	\resizebox{\linewidth}{!}{
		\begin{tabular}{c c c c c c c}
			\toprule
			\multirow{2.4}{*}{Session} & \multirow{2.4}{*}{Method} & \multirow{2.4}{*}{Venue} & \multicolumn{2}{c}{Skilled Forgery $\downarrow$} & \multicolumn{2}{c}{Random Forgery $\downarrow$}\\
			\cmidrule(r){4-5}\cmidrule{6-7}
			~ & ~ & ~ & 4 vs 1 & 1 vs 1 & 4 vs 1 & 1 vs 1\\
			\hline
			\multirow{6}{*}{1} & DTW \cite{vintsyuk1968speech} & - & 48.11/46.34 & 41.82/39.16 & 29.89/28.23 & 36.89/35.26\\
			\cline{2-7}
			~ & DeepDTW \cite{wu2019DeepDTW} & ICDAR'19 & 7.49/3.81 & 11.04/4.01 & 17.50/14.14 & 22.71/13.47\\
			\cline{2-7}
			~ & TA-RNNs \cite{tolosana2021deepsign} & TBIOM'21 & 4.47/1.65 & 7.08/3.49 & 7.86/6.40 & 12.31/9.98\\
			\cline{2-7}
			~ & Sig2Vec \cite{lai2021synsig2vec} & TPAMI'22 & 6.25/3.19 & 9.74/3.91
			& 8.23/6.15 & 12.57/7.41\\
			\cline{2-7}
			~ & DsDTW \cite{jiang2022dsdtw} & TIFS'22 & 5.54/2.71 & 9.05/4.09 & 7.26/5.54 & 11.88/9.32\\
			\cline{2-7}
			~ & PAVENet (\textbf{Ours}) & This work & \textbf{2.67/0.90} & \textbf{5.53/2.07} & \textbf{3.25/1.68} & \textbf{7.51/5.14}\\
			\hline
			
			\multirow{6}{*}{2} & DTW \cite{vintsyuk1968speech} & - & 49.11/47.00 & 42.67/37.01 & 31.21/29.46 & 37.95/35.76\\
			\cline{2-7}
			~ & DeepDTW \cite{wu2019DeepDTW} & ICDAR'19 & 6.41/3.29 & 9.20/3.14 & 15.46/12.39 & 18.37/11.85\\
			\cline{2-7}
			~ & TA-RNNs \cite{tolosana2021deepsign} & TBIOM'21 & 3.15/0.80 & 5.75/2.27 & 6.27/5.01 & 11.62/9.53\\
			\cline{2-7}
			~ & Sig2Vec \cite{lai2021synsig2vec} & TPAMI'22 & 4.99/1.61 & 7.52/2.45 & 7.47/5.40 & 11.52/6.51\\
			\cline{2-7}
			~ & DsDTW \cite{jiang2022dsdtw} & TIFS'22 & 4.19/1.74 & 7.63/2.95 & 7.82/6.09 & 12.84/10.25\\
			\cline{2-7}
			~ & PAVENet (\textbf{Ours}) & This work & \textbf{1.86/0.39} & \textbf{3.37/1.47} & \textbf{2.77/1.26} & \textbf{6.28/3.54}\\
			\hline
			
			\multirow{6}{*}{1 \& 2} & DTW \cite{vintsyuk1968speech} & - & 42.92/41.28 & 42.89/38.62 & 33.45/32.60 & 40.17/38.52\\
			\cline{2-7}
			~ & DeepDTW \cite{wu2019DeepDTW} & ICDAR'19 & 10.32/7.99 & 13.97/8.29 & 26.01/25.74 & 31.56/25.73\\
			\cline{2-7}
			~ & TA-RNNs \cite{tolosana2021deepsign} & TBIOM'21 & 6.77/4.93 & 13.16/8.52 & 21.41/20.24 & 25.53/24.85\\
			\cline{2-7}
			~ & Sig2Vec \cite{lai2021synsig2vec} & TPAMI'22 & 7.58/6.54 & 14.46/7.86 & 16.57/14.51 & 23.68/17.29\\
			\cline{2-7}
			~ & DsDTW \cite{jiang2022dsdtw} & TIFS'22 & 7.32/5.54 & 14.26/10.65 & 13.30/11.77 & 23.86/21.96\\
			\cline{2-7}
			~ & PAVENet (\textbf{Ours}) & This work & \textbf{3.92/2.33} & \textbf{10.15/6.84} & \textbf{6.24/5.15} & \textbf{17.23/13.32}\\
			\noalign{\vspace{-2pt}}
			\bottomrule
	\end{tabular}}
\end{table}

\begin{table}[t]
	\renewcommand{\arraystretch}{1.4}
	\caption{Comparisons with existing methods of verification Equal Error Rates (EER\%) tested on e-BioDigit-L and e-BioDigit-S datasets. Only random forgery can be tested on e-BioDigit-L and e-BioDigit-S since no skilled forgery data is contained in these two datasets.}
	\label{Table::sota-ebio}
	\centering
	\resizebox{\linewidth}{!}{
		\begin{tabular}{c c c c c c c}
			\toprule
			\multirow{3.6}{*}{Session} & \multirow{3.6}{*}{Method} &
			\multirow{3.6}{*}{Venue} &
			\multicolumn{4}{c}{Random Forgery $\downarrow$}\\
			\cmidrule(r){4-7}
			~ & ~ & ~ & \multicolumn{2}{c}{e-BioDigit-L} & \multicolumn{2}{c}{e-BioDigit-S}\\
			\cmidrule(r){4-5}\cmidrule{6-7}
			~ & ~ & ~ & 4 vs 1 & 1 vs 1 & 4 vs 1 & 1 vs 1\\
			\hline
			\multirow{6}{*}{1} & DTW \cite{vintsyuk1968speech} & - & 29.93/28.92 & 35.59/33.93 & 34.11/33.09 & 41.42/40.05\\
			\cline{2-7}
			~ & DeepDTW \cite{wu2019DeepDTW} & ICDAR'19 & 18.69/16.34
			& 21.91/18.12 & 29.78/28.47 & 31.56/28.09\\
			\cline{2-7}
			~ & TA-RNNs \cite{tolosana2021deepsign} & TBIOM'21 & 32.81/32.59 & 36.56/34.83 & 36.96/36.53 & 40.18/37.48\\
			\cline{2-7}
			~ & Sig2Vec \cite{lai2021synsig2vec} & TPAMI'22 & 13.69/10.45 & 20.14/14.87
			& 28.52/25.97 & 33.40/28.25\\
			\cline{2-7}
			~ & DsDTW \cite{jiang2022dsdtw} & TIFS'22 & 17.54/16.80 & 22.31/21.21 & 24.55/22.83 & \textbf{29.61}/26.52\\
			\cline{2-7}
			~ & PAVENet (\textbf{Ours}) & This work & \textbf{10.19/7.69} & \textbf{17.19/12.48} & \textbf{23.33/21.39} & 29.99/\textbf{26.39}\\
			\hline
			
			\multirow{6}{*}{2} & DTW \cite{vintsyuk1968speech} & - & 34.82/32.83 & 39.10/36.50 & 38.58/37.89 & 43.68/43.46\\
			\cline{2-7}
			~ & DeepDTW \cite{wu2019DeepDTW} & ICDAR'19 & 21.63/20.31 & 24.28/18.90 & 31.08/29.09 & 31.72/30.13\\
			\cline{2-7}
			~ & TA-RNNs \cite{tolosana2021deepsign} & TBIOM'21 & 31.41/30.20 & 36.60/33.46 & 36.69/35.92 & 39.79/38.01\\
			\cline{2-7}
			~ & Sig2Vec \cite{lai2021synsig2vec} & TPAMI'22 & 13.14/11.88 & 20.36/14.76 & 33.61/32.58 & 37.39/32.92\\
			\cline{2-7}
			~ & DsDTW \cite{jiang2022dsdtw} & TIFS'22 & 17.01/16.41 & 22.38/21.53 & 25.47/24.39 & 31.44/31.16\\
			\cline{2-7}
			~ & PAVENet (\textbf{Ours}) & This work & \textbf{10.65/9.72} & \textbf{17.71/15.01} & \textbf{23.42/21.36} & \textbf{30.59/27.32}\\
			\hline
			
			\multirow{6}{*}{1 \& 2} & DTW \cite{vintsyuk1968speech} & - & 33.82/32.09 & 39.46/37.85 & 37.04/35.93 & 42.90/40.63\\
			\cline{2-7}
			~ & DeepDTW \cite{wu2019DeepDTW} & ICDAR'19 & 26.41/25.30 & 26.22/24.68 & 32.45/31.51 & 32.75/32.09\\
			\cline{2-7}
			~ & TA-RNNs \cite{tolosana2021deepsign} & TBIOM'21 & 39.23/39.00 & 41.26/40.32 & 42.14/41.26 & 44.21/42.44\\
			\cline{2-7}
			~ & Sig2Vec \cite{lai2021synsig2vec} & TPAMI'22 & 17.03/15.66 & 25.41/19.39 & 32.41/30.97 & 37.49/32.94\\
			\cline{2-7}
			~ & DsDTW \cite{jiang2022dsdtw} & TIFS'22 & 22.65/21.16 & 28.76/27.41 & \textbf{28.36}/28.25 & 34.03/33.15\\
			\cline{2-7}
			~ & PAVENet (\textbf{Ours}) & This work & \textbf{12.42/10.61} & \textbf{21.60/16.95} & 28.55/\textbf{27.32} & \textbf{31.31/27.12}\\
			\noalign{\vspace{-2pt}}
			\bottomrule
	\end{tabular}}
\end{table}

\begin{figure}[t]
	\centering
	\includegraphics[width=\linewidth]{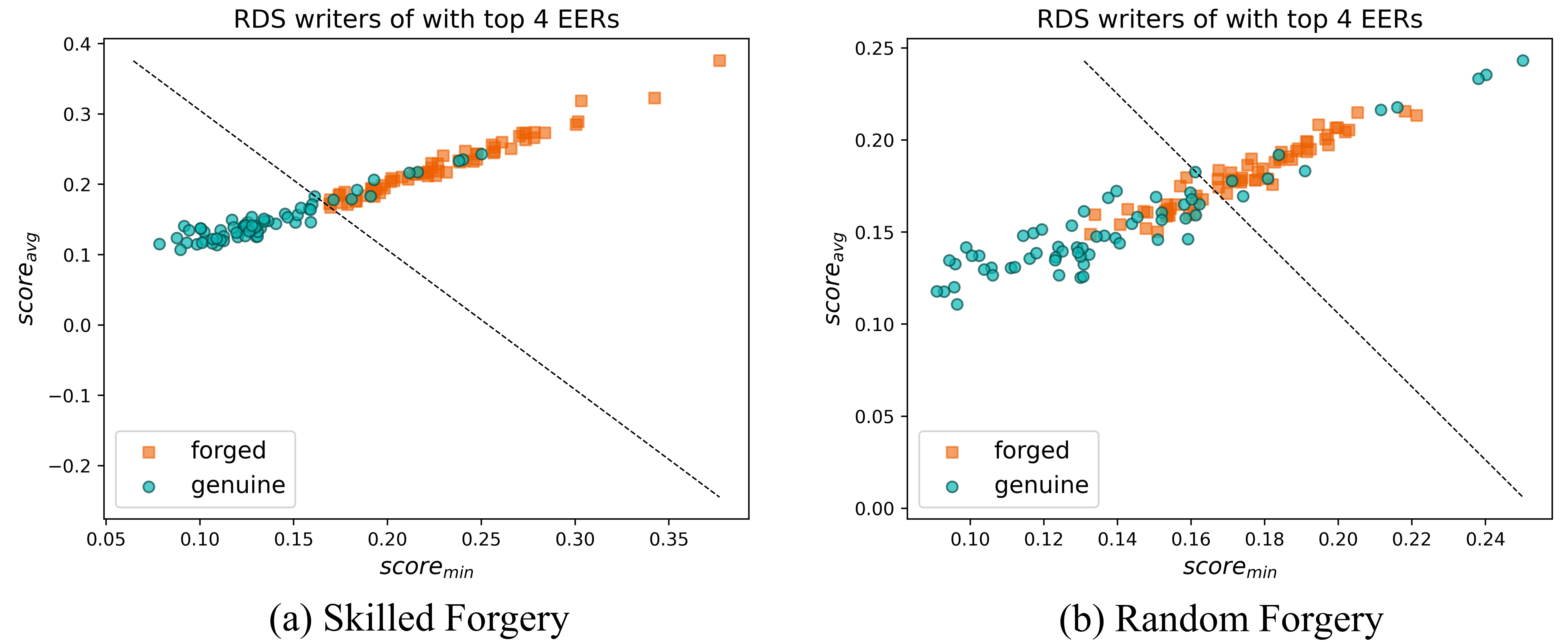}{}
	\caption{Scatter plots of verification results on across-session RDS with the 4 vs 1 template selection strategy. We respectively plot the results of writers with the top 4 highest EERs in skilled forgery and random forgery scenarios.}
	\label{Fig::scatter}
\end{figure}

\subsubsection{Naturalness of the Handwritten RDS}
To scrutinize the significance of the naturalness of handwritten RDS, we conduct experiments on the synthetic datasets e-BioDigit-L/S. To ensure fair comparisons, we structure assessments both on single-session and across-session synthetic data mirroring the evaluation of HRDS4BV. The results are presented in Table~\ref{Table::sota-ebio}. It is important to note that only random forgery can be tested due to the lack of skilled forgery data in these two datasets. In across-session scenarios, the best results on synthetic RDS are merely 12.42/10.61 ($EER_{g}/EER_{l}$) for skilled forgery and 21.60/16.95 for random forgery, significantly inferior to 1.86/0.39 and 3.92/2.33 (Table~\ref{Table::sota-hrds}) achieved on natural RDS. These results underscore the indispensability of the naturalness we have considered in handwritten RDS and the high practicability of our HRDS4BV dataset. In addition, the results indicate a positive correlation between verification accuracy and increased RDS length, as all methods yielded lower EERs on e-BioDigit-L (lengths between 7-11) compared to e-BioDigit-S (lengths between 3-4). Notably, even with synthetic data, PAVENet outperforms other methods by obvious margins in almost all cases. This observation highlights PAVENet's strong generalization capabilities, as it consistently delivers superior performances on both natural and synthetic RDS.

\subsubsection{Inter-Session Variability}
We observe a conspicuous degradation in EERs when systems are tested on across-session RDS compared to single-session RDS. For instance, the single-session EERs of 1.86/0.39 ($EER_{g}/EER_{l}$) for skilled forgery and 2.77/1.26 for random forgery deteriorate to 3.92/2.33 and 6.24/5.15, respectively. The underlying reason for this degradation is the inter-session variability problem, wherein the writer's writing style may exhibit significant variation across different writing sessions due to the impact of one's physical condition, emotional state, writing devices, and other contextual factors. Such variations reasonably reduce handwriting consistency, thereby exacerbating the challenges of writing style learning and, consequently, compromising verification accuracy. In contrast, within a short period (a single session), the writing style remains highly consistent, rendering lower verification difficulty. This phenomenon, which has been reported in existing signature verification studies \cite{kholmatov2009susig,tolosana2019biotouchpass,zhang2022msds}, necessitates serious consideration as it presents challenges for the practical application of authentication systems in terms of stability and reliability.

\subsubsection{Scatter Distribution}
We visualize the verification results of PAVENet on across-session RDS using scatter plots in Fig.~\ref{Fig::scatter}. Points residing below the division line are accepted as genuine samples, while those on the opposing side points are considered forged. For brevity, we only display the results of writers with the top 4 highest EERs. Comparing Fig.~\ref{Fig::scatter} (a) and (b), skilled forgeries exhibit a more centralized distribution, with a majority of samples located on one side of the decision boundary. Conversely, random forgeries exhibit a dispersed location distribution, with a larger number of samples spread on the opposite side of the decision boundary. The misjudgment becomes far more severe in random forgery as there is a greater mixture of genuine and forged samples (Fig.~\ref{Fig::scatter} (b)). Since a more centralized scatter distribution represents better verification accuracy, the scatter distributions indicate that the performance of random forgery lags behind that of skilled forgery. This observation aligns with the numerical results presented in Table~\ref{Table::sota-hrds} and the ROC curves in Fig.~\ref{Fig::roc}, \emph{i.e.}, EERs of random forgery are higher than those of skilled forgery (apart from the untrained DTW method). This is a new and intriguing phenomenon that will be investigated in the subsequent sections.

\subsection{Ablation Study}
\begin{table}[t]
	\renewcommand{\arraystretch}{1.3}
	\caption{Ablation studies on the HRDS4BV. SP denotes that we replace the attentive statistical pooling \cite{okabe2018} with the selective pooling \cite{lai2021synsig2vec}.}
	\label{Table::ablation}
	\centering
	\resizebox{\linewidth}{!}{
		\begin{tabular}{c c c c c c c c c}
			\toprule
			\multirow{2.4}{*}{Baseline} & \multirow{2.4}{*}{SP} & \multirow{2.4}{*}{GTA} & \multirow{2.4}{*}{R-Del} & \multirow{2.4}{*}{DPM} & \multicolumn{2}{c}{Skilled Forgery $\downarrow$} & \multicolumn{2}{c}{Random Forgery $\downarrow$}\\
			\cmidrule(r){6-7}\cmidrule{8-9}
			~ & ~ & ~ & ~ & ~ & 4 vs 1 & 1 vs 1 & 4 vs 1 & 1 vs 1\\
			\hline
			\checkmark & ~ & ~ & ~ & ~ & 6.74/4.45 & 13.41/10.71 & 10.28/9.14 & 21.11/19.02\\
			\hline
			\checkmark & \checkmark & ~ & ~ & ~ & 5.23/3.90 & 12.02/9.02 & 8.33/7.32 & 18.98/15.78\\
			\hline
			\checkmark & \checkmark & \checkmark & ~ & ~ & 4.04/2.68 & 10.55/7.49 & 7.54/6.38 & \textbf{17.20}/13.88\\
			\hline
			\checkmark & \checkmark & \checkmark & ~ & \checkmark & \textbf{3.89/2.30} & 10.35/\textbf{6.74} & 6.53/5.84 & 17.51/14.07\\
			\hline
			\checkmark & \checkmark & \checkmark & \checkmark & ~ & 4.00/2.53 & 10.43/6.96 & 6.97/5.89 & 18.75/14.39\\
			\hline
			\checkmark & \checkmark & \checkmark & \checkmark & \checkmark & 3.92/2.33 & \textbf{10.15}/6.84 & \textbf{6.24/5.15} & 17.23/\textbf{13.32}\\
			\noalign{\vspace{-2pt}}
			\bottomrule
	\end{tabular}}
\end{table}

We conduct ablation studies on the HRDS4BV dataset to examine the effectiveness of proposed model components in a progressive combination manner. Results are listed in Table~\ref{Table::ablation}. The baseline denotes the last time delay scaling block followed by an attentive statistical pooling \cite{okabe2018}. Comparing results in the first line and those of existing methods in Table~\ref{Table::sota-hrds} (Session 1 \& 2), the baseline has already outperformed other methods.

\textbf{Selective Pooling.} Since the PAVENet outputs feature vectors for metric learning-based optimization, we need a powerful technique to transform the time series into fixed-length vectors. We investigate the traditional average pooling, the attentive statistical pooling \cite{okabe2018}, and the selective pooling \cite{lai2021synsig2vec}. Eventually, we choose selective pooling as it presents better performances and employ it as the basic fixed-length vector transformation scheme in all required cases.

\textbf{Global Temporal Attention.} Before adding the Global Temporal Attention (GTA) module, the model lacks long-term dependency modeling. To circumvent this shortcoming, we devise GTA to explicitly introduce global dependency of the entire feature sequence. The final incorporation of style and global features can achieve more complementary handwriting feature representations. With the GTA module, the 4 vs 1 skilled forgery EER is improved by over 1\%. 

\begin{figure}[t]
	\centering
	\includegraphics[width=0.9\linewidth]{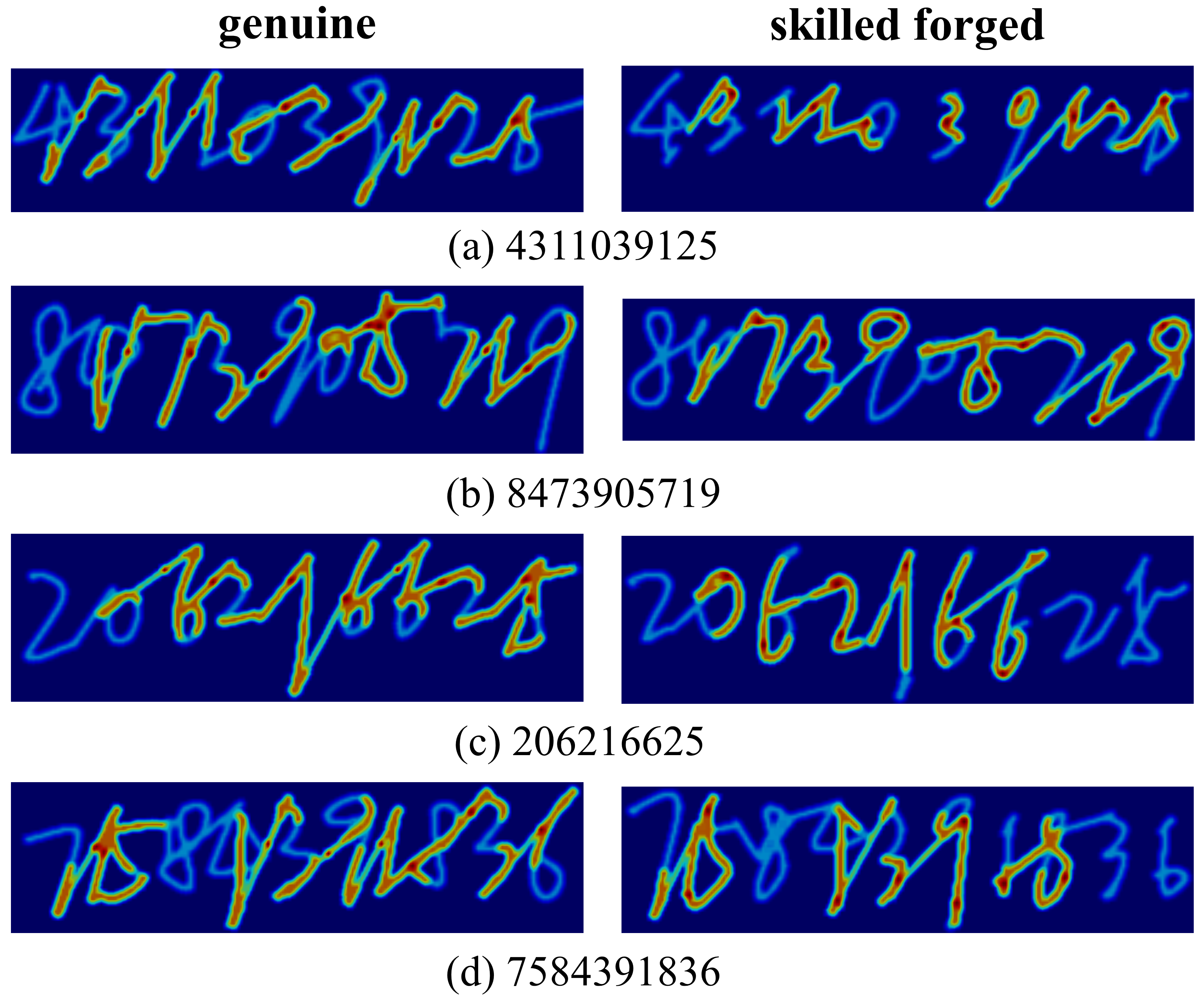}{}
	\caption{Visualization of the pattern key points and pattern segments on RDS detected by the DPM module.}
	\label{Fig::visualize}
\end{figure}

\textbf{Discriminative Pattern Mining.} The efficacy of the Discriminative Pattern Mining (DPM) module is evaluated with/without the use of R-Del. When excluding R-Del, DPM improves random forgery by 1\% and skilled forgery by 0.15\%. With the optimization of R-Del, DPM further improves random forgery by 0.75\% and skilled forgery by 0.08\% and achieves optimal performances. It is observed that DPM achieves higher improvements in random forgery than skilled forgery, which suggests that DPM works better in distinguishing handwriting with completely different styles. This is consistent with the optimization hotspot in RDS verification as improving random forgery is more crucial (discussed in Section~\ref{sec::analysis}). Previous methods are weak in style modeling and yield subpar results due to the absence of content information, while this module enhances style learning by excavating unique and recurring visual writing patterns developed by long-established human writing habits. Therefore, it effectively strengthens personalized handwriting feature representation and boosts model performance.

\subsection{DPM Visualization}
\label{sec::rds_visualize}
Fig.~\ref{Fig::visualize} presents several examples of RDS with handwriting patterns discovered by our proposed DPM module. Each handwriting is highlighted with different pattern key points and segments that reflect their distinctive styles. The most frequently detected patterns are the stroke hyphenations and stroke twirls, which are important writing features commonly observed in naturally handwritten digit strings. Comparing the samples in the two columns, the salient writing patterns detected in genuine and skilled-forged handwriting are poles apart. This facilitates PAVENet to effectively differentiate between genuine RDS and their forged counterparts. The awareness of these discriminative writing patterns not only proves the promising attentive ability of DPM, but also emphasizes the importance of the naturalness embodied in our proposed RDS and HRDS4BV dataset.

\subsection{Analysis of the Computational Resources}
We assess the computational requirement of the proposed PAVENet in terms of running cost and storage cost. For running cost, we test the training and inference time of PAVENet on a 6-core Intel Core i5-8600K CPU alone and four types of GPU, namely NVIDIA TITAN X 12GB, NVIDIA GTX 1080Ti 11GB, NVIDIA RTX 2080Ti 11GB, and NVIDIA RTX 3090 24GB, standing for varying scenarios ranging from limited to sufficient computational resources. Each type of GPU is deployed on one machine with the same 6-core CPU of Intel Core i5-8600K @ 3.60GHz and 64 GB RAM. The running time cost and storage requirement of PAVENet are summarized in Table~\ref{Table::computation}.

As outlined in Sec.~\ref{sec::protocol} and Sec.~\ref{sec::impl_details}, PAVENet is trained on 8,080 samples with a batch size of 18 for 300 epochs, while tested on 8,000 samples. Although the training procedure takes more than three days using merely the CPU, it can be accelerated to 4.75h using RTX 3090, with less than 11GB GPU memory consumption. The inference procedure consumes time ranging from 64.01ms and 1.68ms per sample. Notably, even in the CPU-only scenario, the inference time remains low. GPU acceleration further reduces this to less than 10ms per sample, with the fastest record at 1.68ms per sample on the RTX 3090. Given that human perception considers response times under 100ms as instantaneous, PAVENet's inference speed comfortably meets the needs of real-time applications across various hardware configurations. For storage requirements, PAVENet has 7.65 million (M) parameters, necessitating merely 29.30MB of hard disk storage. Both the low inference time and modest storage consumption demonstrate the computation and storage efficiency of PAVENet.

\begin{table}[t]
	\renewcommand{\arraystretch}{1.2}
	\caption{Running cost and storage cost of PAVENet. The training and test data include 8,080 and 8,000 samples, respectively. ``T.'' denotes Time. ``/s'' denotes per sample.}
	\label{Table::computation}
	\centering
	\resizebox{\linewidth}{!}{
		\begin{tabular}{c c c c| c | c c}
			\toprule
			Hardware & GPU Memory & T. \tiny{Training} & \multicolumn{2}{c|}{T. \tiny{Inference}} & \#Parameters & \#Storage\\
			\midrule
			Intel Core i5-8600K CPU & N/A & 88.03h & 512.08s & 64.01ms/s & \multirow{6.7}{*}{7.65M} & \multirow{6.7}{*}{29.30MB}\\
			\cmidrule{1-5}
			NVIDIA TITAN X & 12GB & 16.43h & 62.82s & 7.85ms/s & ~ & ~\\
			\cmidrule(r){1-5}
			NVIDIA GTX 1080Ti & 11GB & 7.18h & 36.20s & 4.53ms/s & ~ & ~\\
			\cmidrule(r){1-5}
			NVIDIA RTX 2080Ti & 11GB & 6.10h & 26.94s & 3.37ms/s & ~ & ~\\
			\cmidrule(r){1-5}
			NVIDIA RTX 3090 & 24GB & 4.75h & 13.47s & 1.68ms/s & ~ & ~\\
			\bottomrule
	\end{tabular}}
\end{table}

\subsection{Forgery Analysis and New Observation}
\label{sec::analysis}
In signature verification \cite{lai2021synsig2vec,jiang2022dsdtw,zhang2022msds,tolosana2019biotouchpass}, it is widely acknowledged that random forgery is inherently easier verified than skilled forgery owing to the entirely different text content to genuine sample, resulting in lower verification EERs. Nevertheless, all results of the trained models (excluding the untrained DTW method) in Table~\ref{Table::sota-hrds}, Fig.~\ref{Fig::roc}, and Fig.~\ref{Fig::scatter} universally present that the verification EERs of skilled forgery are lower than those of random forgery, which indicates that skilled forgery is easier distinguished, totally in contrast to prior results. The key difference between signature, TDS, and RDS verification lies in whether the biometric trait is text-independent. Therefore, we speculate that this phenomenon is mainly attributed to the text-independent nature of RDS.

Since both RDS and TDS are digit strings but exhibit contrasting text dependencies, we conduct a cross-dataset validation on HRDS4BV and MSDS-TDS \cite{zhang2022msds} datasets using PAVENet to make further investigation, with results presented in Table~\ref{Table::cross}. The cross-dataset validation reveals distinct EER patterns for skilled forgeries ($EER_{sf}$) and random forgeries ($EER_{rf}$) across different training and testing data configurations: (1) \textbf{Using different training data.} Given TDS as fixed test data, the model yields lower $EER_{rf}$ than $EER_{sf}$ (0.38/0.10 versus 2.54/0.66) when trained with TDS but higher $EER_{rf}$ (0.81/0.33 versus 0.69/0.20) with RDS. (2) \textbf{Using different test data.} Given either TDS or RDS as fixed training data, the model yields lower $EER_{rf}$ than $EER_{sf}$ (0.38/0.10 versus 2.54/0.66) when tested with TDS but higher $EER_{rf}$ (28.76/27.73 versus 21.63/20.05) with RDS.

\begin{table}[t]
	\renewcommand{\arraystretch}{1.4}
	\caption{Cross-dataset validation results on the proposed HRDS4BV and MSDS-TDS \cite{zhang2022msds} using PAVENet.}
	\label{Table::cross}
	\centering
	\resizebox{\linewidth}{!}{
		\begin{tabular}{c c c c c c}
			\toprule
			\multirow{2.4}{*}{Training} & \multirow{2.4}{*}{Testing} & \multicolumn{2}{c}{Skilled Forgery $\downarrow$} & \multicolumn{2}{c}{Random Forgery $\downarrow$}\\
			\cmidrule(r){3-4}\cmidrule{5-6}
			~ & ~ & 4 vs 1 & 1 vs 1 & 4 vs 1 & 1 vs 1\\
			\hline
			RDS & RDS & 3.92/2.33 & 10.15/6.84 & 6.24/5.15 & 17.23/13.32\\
			\hline
			RDS & TDS & 0.69/0.20 & 6.25/2.76 & 0.81/0.33 & 8.83/5.15\\
			\hline
			TDS & RDS & 21.63/20.05 & 25.64/23.20 & 28.76/27.73 & 38.14/35.45\\
			\hline
			TDS & TDS & 2.54/0.66 & 8.16/3.19 & 0.38/0.10 & 3.75/1.02\\
			\noalign{\vspace{-2pt}}
			\bottomrule
	\end{tabular}}
\end{table}

These findings suggest that the text-independent property of RDS fundamentally influences both model training and testing, leading to greater difficulties in detecting random forgeries. We attribute this challenge to three key statistical factors: (1) \textbf{Lack of content cues.} Unlike text-dependent handwriting where genuine and forged samples form well-separated feature clusters based on content, RDS samples spread diffusely in the feature space due to the absence of fixed content. Since random forgeries are genuine samples yet from different writers, they exhibit increased overlap in the distribution with authentic samples, impeding model discrimination. (2) \textbf{Writing style deviations.} Skilled forgeries introduce consistent deviations in writing styles that shift their distribution from genuine samples. In contrast, without deliberate imitation, random forgeries lack such telltale deviations and exhibit a more uniform distribution across the style space, rendering them more difficult to distinguish. (3) \textbf{Cognitive and motor control inconsistencies.} While skilled forgeries often exhibit abnormal motor inconsistency due to conscious mimicry attempts, random forgeries maintain natural motor control patterns. Their statistical properties, such as the moments in velocity profiles and stroke variability, more closely approximate those of genuine samples, further complicating their identification.

Furthermore, this novel forgery phenomenon underscores the necessity to prioritize the optimization of random forgery in RDS verification, rather than focusing on improving the classification of skilled forgery in conventional signature verification efforts. Additionally, it suggests that skilled forgeries of RDS could be naturally more difficult to counterfeit than those of other biometric forms (\emph{e.g.}, TDS and signature), due to the challenge that skilled forgers encounter in accurately replicating diverse handwriting styles presented in variable content. From the perspective of anti-counterfeiting, this discovery could have positive implications and inspire advancements in defending against skilled forgery.

\section{Discussion}
\label{sec::discussion}

\subsection{Application}
The inherent uniqueness of each individual's personal handwriting style, originating from a combination of physiological and behavioral characteristics, such as finger dexterity, muscle memory, and writing habits, presents a compelling advantage for leveraging handwriting as a biometric. Consequently, personal handwriting has found applications in various fields for decades. In forensic handwriting examination \cite{strengtheningUSAapath2009national,malik2014man}, experts scrutinize signatures and handwritten notes to verify their authenticity, providing crucial support to judicial processes. In e-commerce applications \cite{commercial1997signature,commercialaspect1999,diaz2019aperspective}, users authenticate themselves through online signatures to ensure transaction safety, spurring the development of considerable commercial signature verification products that provide customized and confidential services to customers. Meanwhile, in mobile authentications, some studies have explored the use of touchscreen handwriting biometrics such as finger-drawn Person Identification Numbers (PIN) \cite{nguyen2017draw} and One Time Passwords (OTP) \cite{tolosana2019biotouchpass,tolosana2020biotouchpass2} as enhancements to standard typed passwords. They employed handwritten single digits other than signatures and proved the efficacy of handwritten digits in enhancing traditional type-based password/verification code authentication systems.

Nevertheless, no prior research has explored how to tackle the privacy violation issue in identity verification, nor has any attempt considered utilizing handwritten RDS as a biometric medium for identity verification. Indeed, the privacy-protective nature of RDS presents a crucial advantage over other forms of handwriting, making it a highly promising and valuable means for verification. This feature is innately beneficial in sensitive scenarios like online financial transactions and banking identity confirmation, where RDS can substitute conventional biometrics like signature to avoid inadvertent private information exposure and potential privacy compromises. In addition, the scope of RDS extends beyond a mere combination of randomly generated numbers as employed in our dataset. It can be content-variable digit strings with realistic values, such as contract numbers, order numbers, and dates, enhancing its practicality for acquisition and deployment in diverse contexts. Furthermore, the integration of online RDS with our proposed PAVENet can yield a verification system with promising accuracy. Such a system holds substantial potential for information tracking, enabling the identification and tracing of criminals through verifying their handwriting, such as handwritten dates or hotel order numbers. Additionally, it could be employed as a form of handwriting evidence in forensic investigations, offering critical insights into the identities of individuals involved in criminal activities.

\begin{table}[t]
	\renewcommand{\arraystretch}{1.3}
	\caption{Comparisons between different biometrics, including Spanish signature from MCYT \cite{ortega2003mcyt}, Chinese signature and TDS from MSDS \cite{zhang2022msds}, and RDS from HRDS4BV, using state-of-the-art verification methods.}
	\label{Table::model_biometric_comp}
	\centering
	\resizebox{\linewidth}{!}{
		\begin{tabular}{c c c c c c c}
			\toprule
			\multirow{2.4}{*}{Biometric} & \multirow{2.4}{*}{Method} & 
			\multirow{2.4}{*}{Venue} &
			\multicolumn{2}{c}{Skilled Forgery $\downarrow$} & \multicolumn{2}{c}{Random Forgery $\downarrow$}\\
			\cmidrule(r){4-5}\cmidrule{6-7}
			~ & ~ & ~ & 4 vs 1 & 1 vs 1 & 4 vs 1 & 1 vs 1\\
			\hline
			
			\multirow{6}{*}{\makecell[c]{Spanish\\Signature}} & DTW \cite{vintsyuk1968speech} & - & 2.91/1.77 & 6.20/2.35 & 0.67/0.27 & 1.20/0.39\\
			\cline{2-7}
			~ & DeepDTW \cite{wu2019DeepDTW} & ICDAR'19 & 2.53/1.32 & 4.44/1.67 & 0.46/0.21 & 1.12/0.30\\
			\cline{2-7}
			~ & TA-RNNs \cite{tolosana2021deepsign} & TBIOM'21 & 4.30/- & 4.40/- & 0.20/- & 1.10/-\\
			\cline{2-7}
			~ & Sig2Vec \cite{lai2021synsig2vec} & TPAMI'22 & 2.38/\textbf{0.96} & 3.84/1.59 & \textbf{0.21/0.06} & \textbf{0.55/0.12}\\
			\cline{2-7}
			~ & DsDTW \cite{jiang2022dsdtw} & TIFS'22 & \textbf{1.86}/1.03 & \textbf{3.77/1.53} & 0.49/0.16 & 1.10/0.28\\
			\cline{2-7}
			~ & PAVENet (\textbf{Ours}) & This work & 2.30/1.05 & 10.43/2.74 & 3.11/1.53 & 5.02/1.42\\
			\hline
			
			\multirow{6}{*}{\makecell[c]{Chinese\\Signature}}
			~ & DTW \cite{vintsyuk1968speech} & - & 5.41/2.12 & 19.01/8.70 & \textbf{0.00/0.00} & \textbf{1.03/0.03}\\
			\cline{2-7}
			~ & DeepDTW \cite{wu2019DeepDTW} & ICDAR'19 & 4.17/1.92 & 10.86/3.51 & 1.11/0.37 & 6.49/1.50\\
			\cline{2-7}
			~ & TA-RNNs \cite{tolosana2021deepsign} & TBIOM'21 & 3.41/1.38 & \textbf{8.64/4.13} & 1.36/1.06 & 1.60/0.57\\
			\cline{2-7}
			~ & Sig2Vec \cite{lai2021synsig2vec} & TPAMI'22 & 2.97/0.74 & 12.15/4.11 & 0.19/0.09 & 4.28/0.86\\
			\cline{2-7}
			~ & DsDTW \cite{jiang2022dsdtw} & TIFS'22 & \textbf{1.60/0.57} & 9.66/3.71 & 0.05/0.00 & 1.70/0.34\\
			\cline{2-7}
			~ & PAVENet (\textbf{Ours}) & This work & 2.90/1.27 & 11.41/6.66 & 0.71/0.30 & 7.17/2.47\\
			\hline
			
			\multirow{6}{*}{TDS} & DTW \cite{vintsyuk1968speech} & - & 4.84/1.78 & 17.52/7.64 & \textbf{0.03/0.00} & \textbf{0.36/0.01}\\
			\cline{2-7}
			~ & DeepDTW \cite{wu2019DeepDTW} & ICDAR'19 & 2.26/0.76 & 7.25/0.77 & 0.40/0.11 & 3.18/0.14\\
			\cline{2-7}
			~ & TA-RNNs \cite{tolosana2021deepsign} & TBIOM'21 & 1.73/0.79 & 4.32/1.57 & 1.20/0.89 & 0.93/0.33\\
			\cline{2-7}
			~ & Sig2Vec \cite{lai2021synsig2vec} & TPAMI'22 & \textbf{0.79}/0.21 & 5.23/1.77 & \textbf{0.03}/0.01 & 0.88/0.08\\
			\cline{2-7}
			~ & DsDTW \cite{jiang2022dsdtw} & TIFS'22 & \textbf{0.79/0.20} & 6.18/1.88 & \textbf{0.03/0.00} & 1.10/0.24\\
			\cline{2-7}
			~ & PAVENet (\textbf{Ours}) & This work & 2.54/0.66 & 8.16/3.19 & 0.38/0.10 & 3.75/1.02\\
			\hline
			
			\multirow{6}{*}{RDS} & DTW \cite{vintsyuk1968speech} & - & 42.92/41.28 & 48.89/38.62 & 33.45/32.60 & 40.17/38.52\\
			\cline{2-7}
			~ & DeepDTW \cite{wu2019DeepDTW} & ICDAR'19 & 10.32/7.99 & 13.97/8.29 & 26.01/25.74 & 26.04/24.31\\
			\cline{2-7}
			~ & TA-RNNs \cite{tolosana2021deepsign} & TBIOM'21 & 6.77/4.93 & 13.16/8.52 & 21.41/20.24 & 25.53/24.85\\
			\cline{2-7}
			~ & Sig2Vec \cite{lai2021synsig2vec} & TPAMI'22 & 7.58/6.54 & 14.46/7.86 & 16.57/14.51 & 23.68/17.29\\
			\cline{2-7}
			~ & DsDTW \cite{jiang2022dsdtw} & TIFS'22 & 7.32/5.54 & 14.26/10.65 & 13.30/11.77 & 23.86/21.96\\
			\cline{2-7}
			~ & PAVENet (\textbf{Ours}) & This work & \textbf{3.92/2.33} & \textbf{10.15/6.84} & \textbf{6.24/5.15} & \textbf{17.23/13.32}\\
			\noalign{\vspace{-2pt}}
			\bottomrule
	\end{tabular}}
\end{table}

\subsection{Challenge}
\label{sec::challenge}
Apart from the advantages, this study has left several limitations unaddressed. First, although the text-independent property of RDS benefits personal privacy protection, its verification performance lags behind other biometrics. We compare RDS against three additional types of biometric modalities: Spanish signature, Chinese signature, and Token Digit String (TDS). Spanish signature data is sourced from the MCYT dataset \cite{ortega2003mcyt}, while Chinese signature and TDS originate from the across-session data of MSDS-ChS \cite{zhang2022msds} and MSDS-TDS \cite{zhang2022msds} datasets. These datasets possess equal (MSDS-ChS and MSDS-TDS) or similar (MCYT) sizes to the proposed HRDS4BV, which guarantees fair comparisons. Experimental results are summarized in Table~\ref{Table::model_biometric_comp} and we draw the following observations. (1) On RDS, PAVENet delivers significantly better outcomes than other methods. However, on the other biometric traits, it merely presents comparable performance to other SOTA methods but fails to achieve optimal performance. Performance in several cases is far from satisfactory, such as the random forgery EERs on Chinese signature and Spanish signature. The best performances are typically held by the methods designed for text-dependent verification, such as DTW \cite{vintsyuk1968speech}, DsDTW \cite{jiang2022dsdtw}, and Sig2Vec \cite{lai2021synsig2vec}. This suggests that PAVENet could be a specialist for text-independent handwriting verification, such as RDS verification, while somewhat lacking versatility in text-dependent verification tasks, such as signature verification. (2) The general performance of RDS verification lags behind that of signature/TDS verification. Comparing the optimal outcomes between the other three biometrics and RDS, \emph{i.e.}, 1.86/0.96, 0.21/0.06 on Spanish signature, 1.60/0.57 and 0.00/0.00 on Chinese signatures, 0.79/0.20 and 0.03/0.00 on TDS, respectively, they are consistently better than 3.92/2.33 and 6.24/5.15 on RDS. This performance discrepancy can be attributed to the higher verification difficulty introduced by the unconstrained content (text-independent property) and writing style variation of RDS, which increases the likelihood of system misjudgment under the forgery attacks and damages system reliability. From these two findings, although our PAVENet largely outperforms existing SOTA on RDS verification, additional efforts for more powerful systems on this task are still required to meet or surpass the performance bar set by other biometrics. However, it's important to underscore that the privacy-preserving nature of RDS is an inherent and non-negligible merit that other handwriting biometrics lack. This unique feature should be prioritized and considered when assessing the overall suitability of RDS for applications.

Second, it is worth noting that the size of the collected HRDS4BV dataset is somewhat limited. Nonetheless, it's crucial to acknowledge the significant challenges associated with the data collection process for such a dataset, which is largely due to the requirement for on-site data acquisition that necessitates dedicated guidance and supervision. The effort, time, equipment, and cost needed to acquire, correct, and cleanse a dataset of such a size are enormous. In fact, the size of the HRDS4BV dataset is comparable to or exceeds the size of most public existing handwriting verification datasets \cite{lu2017scut,tolosana2019biotouchpass,zhang2022msds}. Still, from the academic perspective, collecting more online RDS data from a broader range of writers could yield more compelling experimental results.

Third, we have not fully addressed the inter-session variability problem, which refers to the inconsistency in writing features induced over time. Owing to the significant time and labor cost of data curation, we opted to conduct data collection across only two acquisition sessions, with a time gap of just 21 days. In practice, however, the time gap between users' different enrolments to the system could span months or even years, and users may input handwriting to the same system more than two times. Hence, it would be beneficial to extend the acquisition time span as writers' writing styles may exhibit more variations over longer periods. Increasing the amount of collection sessions could also more closely approach realistic scenarios and facilitate studying long-term writing style variability. Despite these considerations, it's important to note that our system is not specifically engineered to address the inter-session variability problem. Exploring more advanced modeling schemes for time-invariant feature extraction could be instrumental in addressing this variability and we expect to delve into these techniques in our future work.

\subsection{Ethical Consideration}
As we intend to publish the HRDS4BV dataset, we have taken ethical considerations and made efforts to ensure the proper use of the data as well as protect the rights of writers. First, before data collection, we obtained informed consent from the writers by having each of them sign a copyright agreement. This agreement grants us the license to use their handwriting data for academic research and publication. The copyright agreements are included in Fig. A, Appendix A. Second, to avoid potential misuse and negative societal impact, we have meticulously defined the terms of use for our dataset, which primarily encompass the following three rules. (1) The HRDS4BV dataset is strictly and exclusively intended for non-commercial research purposes. Researchers intending to use this dataset are required to undergo a verification process through the application procedure outlined in our official GitHub page. (2) While the HRDS4BV dataset contains the same number of writers as the MSDS, these writers are not affiliated with those in the MSDS. (3) We have provided a legally binding commitment that users must sign to prevent any illegal use of the data. In addition, data contributors have the right to withdraw (remove) their handwritten data from the HRDS4BV dataset.

\subsection{Future Exploration}
Several directions are well worth exploring in future work. First, the HRDS4BV dataset could be further extended, such as (1) collecting Random Digit String (RDS) data across more acquisition sessions and over a longer acquisition time interval to thoroughly investigate the inter-session variability, and (2) enlarging the dataset size by collecting data from more writers to enable more robust evaluations of RDS. Second, exploring new handwritten biometric mediums for handwriting verification presents promising opportunities. This direction includes (3) expanding handwriting to various languages and scripts, (4) incorporating different levels of handwriting difficulty, ranging from simple numerals to complex alphanumeric strings, and (5) investigating other biometric modalities that contain unique writing characteristics, such as the audio signals of pen stroke \cite{mmhsv2024icassp} or hand movement videos during writing. Moreover, as outlined in Sec.~\ref{sec::challenge}, (6) developing more powerful mechanisms to enhance the RDS style modeling could help enhance the performance and applicability of RDS verification.

\section{Conclusion}
\label{sec::conclusion}
In this paper, we propose to utilize online handwritten Random Digit String (RDS) as a novel biometric trait for privacy-preserving online handwriting verification. RDS comprises random and arbitrary digits without including any private information, which is inherently privacy-preserving. This nature allows individuals to verify their identity while upholding their privacy by simply writing a random sequence of numbers. For benchmark evaluation, we specifically build an online RDS dataset named HRDS4BV, consisting of naturally and coherently handwritten RDS. To handle the challenging writing style modeling caused by content variations of unconstrained RDS, we propose a model termed PAVENet, along with a specific DPM module for handwriting pattern mining. DPM functions similarly to the human visual perception of handwritten text, which adaptively extracts the consistent and discriminative handwriting patterns from a high-dimensional perspective, thus fundamentally enhancing writing style representation.

Through quantitive and qualitative experiments, we demonstrate the feasibility of applying natural RDS in handwriting verification and show that the proposed PAVENet prominently outperforms prevailing online verification methods. We surprisingly discover an opposite phenomenon to previous results that skilled forgery is easier verified than random forgery in RDS verification and discuss its potential positive impacts on preventing malicious forgery attacks. Based on our model, RDS holds significant promise for further explorations and applications of privacy-preserving handwriting verification. Its unique and valuable nature, particularly its emphasis on privacy preservation, sets it apart from existing handwriting biometric techniques. We hope the successful implementation and evaluation of RDS in this study could serve as inspirations for future research endeavors in this domain.

\section*{Acknowledgments}
This research is supported in part by National Key Research and Development Program of China (2022YFC3301703) and National Natural Science Foundation of China (Grant No.: 62476093, 62441604).

\bibliographystyle{IEEEtran}
\bibliography{reference} 

\appendices
\section{Writer Copyright Agreement}
\label{appendix::copyright}
\begin{figure}[h]
	\centering
	\includegraphics[width=\linewidth]{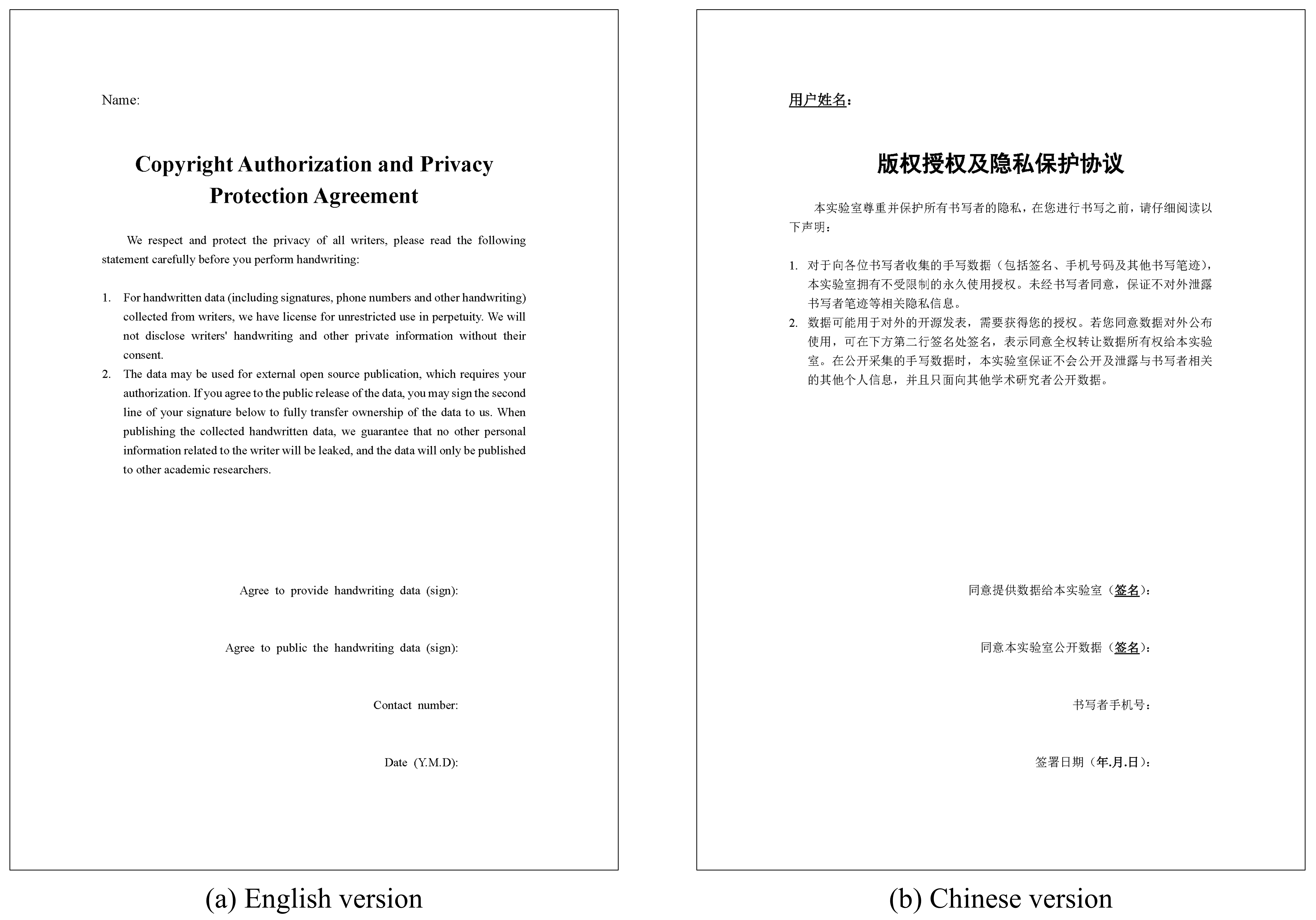}{}
	\caption{Copyright agreement for writer consent.}
	\label{Fig::agreement}
\end{figure}

Fig.~\ref{Fig::agreement} presents the copyright agreements obtained for user consent prior to data collection. The agreements are shown in both the English and Chinese versions.

\end{document}